\documentclass[]{interact}

\usepackage{epstopdf}% To incorporate .eps illustrations using PDFLaTeX, etc.
\usepackage{subcaption}
\usepackage{graphicx}
\usepackage{url}
\usepackage{multirow} 
\usepackage{booktabs}
\usepackage{fancybox}
\usepackage{graphicx}
\usepackage{tcolorbox}
\usepackage{hyperref}
\usepackage{float} 
\usepackage{algorithm}
\usepackage{algpseudocode}

\usepackage{natbib}% Citation support using natbib.sty
\bibpunct[, ]{(}{)}{,}{a}{}{,}% Citation support using natbib.sty
\renewcommand\bibfont{\fontsize{10}{12}\selectfont}% Bibliography support using natbib.sty
\renewcommand\bibfont{\fontsize{10}{12}\selectfont}% Bibliography support using natbib.sty

\theoremstyle{plain}% Theorem-like structures

\theoremstyle{definition}

\theoremstyle{remark}

\usepackage{xcolor}

\begin{document}
\articletype{Research Article}

\title{Georeferencing complex relative locality descriptions with large language models}

\author{
\name{Aneesha Fernando\textsuperscript{a}\textsuperscript{*}\thanks{CONTACT Aneesha Fernando. Email: afwboosa@massey.ac.nz}, Surangika Ranathunga\textsuperscript{a}, Kristin Stock\textsuperscript{a}, Raj Prasanna\textsuperscript{a} and Christopher B. Jones\textsuperscript{b}}
\affil{\textsuperscript{a}Massey University, New Zealand  \textsuperscript{b}Cardiff University, United Kingdom}
}
\maketitle

\begin{abstract}
Georeferencing text documents has typically relied on either gazetteer-based methods to assign geographic coordinates to place names, or on language modelling approaches that associate textual terms with geographic locations. However, many location descriptions specify positions relatively with spatial relationships, making geocoding based solely on place names or geo-indicative words inaccurate. This issue frequently arises in biological specimen collection records, where locations are often described through narratives rather than coordinates if they pre-date GPS. Accurate georeferencing is vital for biodiversity studies, yet the process remains labour-intensive, leading to a demand for automated georeferencing solutions. This paper explores the potential of Large Language Models (LLMs) to georeference complex locality descriptions automatically, focusing on the biodiversity collections domain. We first identified effective prompting patterns, then fine-tuned an LLM using Quantized Low-Rank Adaptation (QLoRA) on biodiversity datasets from multiple regions and languages. Our approach outperforms existing baselines with an average, across datasets, of 65\% of records within a 10 km radius, for a fixed amount of training data. The best results (New York state) were 85\% within 10km and 67\% within 1km. The selected LLM performs well for lengthy, complex descriptions, highlighting its potential for georeferencing intricate locality descriptions. 
\end{abstract}


\section{Introduction}
While most geographical information systems are predominantly based on structured digital map data, there remains a vast amount of information embedded within textual resources. Access to the information content of such resources is the subject of the field of geographical information retrieval \citep{purves_geographic_2018}, which depends to a large extent on the effectiveness of georeferencing methods to determine the geospatial focus of the content of text documents. To date, georeferencing methods for textual data have usually been applied to the content of web pages and of social media postings. The methods typically employ either gazetteer-based approaches to detect and geocode (determine coordinates for) place names in the texts, or language modelling approaches that depend upon determining the association between text and locations, or some combination of these \citep{gritta_whats_2018, melo_automated_2017}. References to geographic locations through the use of place names are usually assumed to be absolute, in the sense that the location is regarded as equivalent to that of a place name or some other words that are indicative of a location \citep{han-etal-2012-geolocation}. Little attention has been given to the fact that some descriptions of location are relative, in that they refer to a location that has some spatial relationship to a reference place name \citep{wieczorek_point-radius_2004, van_erp_georeferencing_2015, chen_georeferencing_2018}. Spatial relational terms include phrases and words that indicate a specified distance, such as \emph{10km west of}, as well as relative positions like \emph{near}, \emph{adjacent to}, and \emph{along from}.

A domain in which complex locality descriptions are commonly found is that of the collection records of natural history agencies such as museums and herbaria. These collections can include records of biological specimens of plants and animals, fungi and bacteria, as well as soil and geological samples. Many of the records, specially those collected before the widespread availability of GPS, do not have associated geographical coordinates. Instead, their locations are described solely through textual descriptions that include locality descriptions. At its simplest, a locality description might just consist of place names (toponyms), but very commonly, place names are combined with relative spatial terms \citep{van_erp_georeferencing_2015, wieczorek_point-radius_2004, scott_automated_2021}. There are billions of such records and there is a strong motivation to georeference them, as being able to assign coordinates, and hence map the locations at which they were found, is a crucial step in studying biodiversity. Such georeferenced data enables researchers to monitor the geographical distribution of species over time, the impacts of environmental changes on species, and to predict how environmental changes will affect biodiversity in specific regions \citep{van_erp_georeferencing_2015}.

To georeference textual documents such as Wikipedia articles, news articles, and social media posts with coordinates that best represent the content, various methods are employed \citep{stock2018mining}. While gazetteer-based approaches are commonly used in this context to detect the presence of place names \citep{karimzadeh2019geotxt}, other methods, such as probabilistic language modelling techniques and discriminative classifiers, have also been utilised. Probabilistic language modelling techniques use generative models that estimate the likelihood of a textual document belonging to a particular region. Discriminative classifiers, such as logistic regression and multi-layer neural networks, classify documents into their most likely regions \citep{melo_automated_2017}. Furthermore, the introduction of Transformer models \citep{vaswani2017attention} has spurred several research efforts using BERT-based models for tweet or tweet user geolocation prediction~\citep{scherrer_heljuvardial_2020, simanjuntak2022we, lutsai2023geolocation, li2023transformer}. 

The often complex relative descriptions of localities where biological specimens are found present an additional challenge from standard document georeferencing approaches mentioned above. As an example, manually georeferencing the locality `\emph{C. 10 km N of Lake Wanaka, 1 km N of Makarora. near Pipson Creek}', which explains the location of a fungi specimen found in the Canterbury district of New Zealand, requires determining the coordinates of Lake Wanaka and the town of Makarora; interpreting the spatial relationship described by `\emph{approximately 10 km north of Lake Wanaka}' and `\emph{1 km north of Makarora}' to find the general vicinity; and finally, using the reference to `\emph{near Pipson Creek}' to pinpoint the precise location.  Thus there is a need for more specialised automated georeferencing techniques. Additional examples of such descriptions are presented in Table~\ref{tab:ex-locality}.

Recently, the field of Natural Language Processing (NLP) has been transformed by the advent of Large Language Models (LLMs), demonstrating advanced functionalities and effective solutions. The vast size of LLMs, which typically include billions of parameters, enables them to discern complex patterns and subtleties in human language, resulting in outputs that are both highly precise and contextually appropriate \citep{fujiwara_modify_2024}.  Leveraging these abilities, LLMs have been successfully applied in various complex domains such as mathematical reasoning \citep{imani_mathprompter_2023, yu_alert_2023} and geospatial reasoning \citep{li_geolm_2023}. They have often been found to outperform previous solutions based on traditional Machine Learning (ML) techniques such as Logistic Regression and BERT-based models \citep{karanikolas2023large}.

In this study, we introduce a novel approach for automatic georeferencing of biological collection data using LLMs. To the best of our knowledge, our study is the first to employ LLMs to georeference text data by predicting coordinates that best match the given locality descriptions. We conducted multiple initial experiments to identify the most suitable LLM and the prompting pattern for our use case. Subsequently, we performed supervised fine-tuning with quantized Low-Rank Adaptation (QLoRA) to adapt the selected LLM for the task of georeferencing. This was applied to multiple datasets containing data from different regions at varying granularities of area size. We also experimented with non-English data to demonstrate the model's linguistic generalizability. The results were benchmarked against several existing georeferencing baselines as well as commercial LLMs. In addition, we investigated the model’s sensitivity to spatial information embedded in locality descriptions and explored its transfer learning capabilities across geographic regions.

To summarize, the contributions of this research are as follows:
\begin{itemize}
    \item We demonstrate the effectiveness of applying LLMs to georeferencing locality descriptions with relative spatial relationships, and establish  a strong benchmark that significantly outperforms the commonly used baselines.
    \item We investigate the optimal LLM prompting patterns for georeferencing use cases, and introduce a new prompt that yields very good results for the considered task.
    \item We analyse the regional and linguistic generalizability of our LLM-based approach for the task of georeferencing. 
    \item We demonstrate that our proposed model is more robust for non-English datasets compared to the baselines. For smaller regions, such as New Zealand, our model, fine-tuned with just 5,000 samples, outperforms the baselines.
\end{itemize}

The remainder of this article is organized as follows. Section 2 describes related work. Section 3 describes the data we used and our methodology for automatic georeferencing. Section 4 discusses the results, and Section 5 concludes the paper with a look into future directions.

\section{Related Work}
\subsection{Georeferencing Text Data}

The idea of georeferencing text documents or specific descriptions of localities and incidents is quite well established \citep{goodchild2004georeferencing, murphy_georeferencing_2004, doherty_georeferencing_2011, melo_automated_2017}. Gazetteer-based methods are commonly used to georeference text by detecting place names~\citep{karimzadeh2019geotxt}. In contrast, other techniques that consider all text, not just the place names, have also been applied, including probabilistic language models, discriminative classifiers, and hybrid approaches that combine these models with gazetteers \citep{melo_automated_2017}.

Gazetteer-based approaches \citep{karimzadeh2019geotxt} primarily focus on geoparsing, and have two steps; toponym recognition, to detect the presence of a location reference, and toponym resolution to disambiguate and hence determine coordinates \citep{wang_are_2019, liu_geoparsing_2022}. Toponym recognition is mostly performed as a Named Entity Recognition (NER) task and the identified toponyms may then be resolved via gazetteer matching in association with a disambiguation process \citep{leidner2003grounding, li2003infoxtract}. However, gazetteers can sometimes be incomplete or outdated, particularly when dealing with historical place names, cultural variations, or locations that have undergone name changes over time, posing challenges for georeferencing based solely on them \citep{sharma_spatially-aware_2023}. To address such limitations, some studies have proposed enhancements, for example, \citet{lieberman2010geotagging} introduced audience-specific local lexicons to better capture regionally relevant toponyms that might be overlooked by global gazetteers.

For the task of georeferencing textual content such as tweets and Wikipedia articles, early research has predominantly utilised ML models, approaching georeferencing as a classification problem. This involves classifying location descriptions at the country and city levels. Probabilistic classifiers, such as n-gram models \citep{flatow2015accuracy, iso2017density} and discriminative classifiers (e.g., ~Support Vector Machines (SVM), and Gaussian Mixture Models (GMM)), have been employed in these studies \citep{melo2015geocoding, priedhorsky2014inferring, liu2015estimating}. In addition to employing classification models, some studies have incorporated clustering techniques and feature selection strategies in combination with various models to georeference documents \citep{laerajones2014, han2014text}. For example, \cite{di2021sherloc} introduced Sherloc, a knowledge-based approach for subcity-level geolocation of tweets. Sherloc improves geolocation accuracy within a specified parent region by matching tweets to a geographic embedding constructed from known toponyms for that region, using clustering approaches to enhance precision.

More recent advancements in ML techniques have introduced Deep Learning models, such as Deep Neural Networks (DNN), Long Short-Term Memory (LSTM) networks and Transformer-based models, for the task of georeferencing. Specifically, georeferencing approaches utilizing pre-trained Transformer models such as BERT have demonstrated stronger performance compared to earlier ML systems. For instance, in the study by \cite{simanjuntak2022we}, BERT significantly outperformed LSTM in predicting Twitter users' home locations. However, their most effective method involved aggregating all tweets from a single user to predict their location, a strategy that is potentially constrained by BERT's 512-token limit.
\cite{scherrer_heljuvardial_2020} demonstrated that BERT in regression mode can be effectively fine-tuned to predict geolocations from text. \cite{EdwardsHybrid2025} built on the latter BERT-based method to georeference social media posts in a hybrid approach, in which the inferred location guided disambiguation of place names when present. Coordinates of the finest-grained place name then represented the tweet’s location, otherwise the transformer-predicted location was used. \cite{lutsai2023geolocation} integrated BERT with two-dimensional GMMs to estimate locations as coordinate pairs. Nevertheless, in addition to adapting pre-trained language models to tweet content, most tweet georeferencing methods also incorporate tweet metadata and user network information as inputs for the model \citep{lutsai2023geolocation, do2018twitter, miura2017unifying}. 

When dealing with generic documents and social media content georeferencing, spatial relation terms are not frequently found, unlike in biological collection records. Equally, the methods mentioned above do not pay attention to spatial relationship terms that provide critical context by describing how one location is positioned relative to another. Such spatial relation terms can include prepositional phrases, verbs, and adverbs \citep{liao_predicting_2022}. Recent work by \cite{li2021neural} proposes a geospatial semantic graph representation to better capture spatial dependencies in text, emphasising that modelling such relationships improves the interpretation of complex geographic descriptions. \cite{moncla2014geocoding} also discuss the importance of spatial relations in estimating the spatial footprint of non-gazetteered place names. In contrast, their geocoding method relies heavily on the clustering of nearby known toponyms, which limits the potential to fully exploit spatial cues provided in the text. 

\cite{chen_georeferencing_2018} have successfully leveraged spatial relations in georeferencing text documents. They convert raw text descriptions to place graphs as the model input and leverage spatial relation models for approximate locating and matching.  However, this methodology is challenging for georeferencing larger datasets with multi-clausal location descriptions as it assumes the prior existence of a place graph, the creation of which could be a complex task in itself. 

This highlights that current georeferencing methods are not well-suited for text in which locations are described in relative terms rather than by absolute place names. These complex descriptions are common in the biological specimen domain, and hence there is a need for more specialised methods tailored to this field, capable of handling complex relative locality descriptions. 
 
\label{Biological specimen georeferencing}
\subsection{Georeferencing Biological Specimen Data}
Among digitised biological specimen records, only a small portion of the data is georeferenced, whereas most records include a verbal locality description
%indicating the location referencing to a place name
\citep{stock_biowhere_2023}.  Several methods have been developed for georeferencing these locality descriptions, including the point-radius method \citep{wieczorek_point-radius_2004}, shape method, bounding box method, and probability method \citep{guo_georeferencing_2008}. These methods involve the time-consuming process of determining precise locations and calculating the associated uncertainty using GIS tools, maps, and aerial photography \citep{review_georeferencing}. \cite{doherty_georeferencing_2011} indicate that the processing time for the point-radius method ranges from 5 to 15 minutes, while the shape method can take between 15 to 90 minutes to georeference a single locality description. Given the sheer volume of records (billions worldwide), these semi-automated methods are not scalable or efficient.

Among studies to automate the georeferencing of biological specimen data, \cite{van_erp_georeferencing_2015} developed a knowledge-driven, rule-based approach. Their method also produces a confidence score to indicate the certainty of the results. \cite{scott_automated_2021} introduced an automated georeferencing method for Antarctic species by developing a pipeline that extracts and processes text from legacy documents. This method identifies species and toponyms within the documents and predicts species-toponym pairs representing actual geospatial relationships. It combines rules and dictionary-based species extraction with place name extraction methods, followed by tree-based classifiers to match species with their toponyms. 

BioGeomancer \citep{guralnick_biogeomancer_2006} and GEOLocate\footnote{https://www.geo-locate.org} are applications developed to automatically georeference biological specimen data. BioGeomancer processes text, interprets it, queries gazetteers, and intersects spatial descriptions, ultimately delivering a standardized geographical reference complete with levels of uncertainty \citep{van_erp_georeferencing_2015}. Unfortunately, at the time of writing, it appears that the BioGeomancer application is no longer functional, and its website is inaccessible.

On the other hand, GEOLocate is a fully operational stand-alone online application that translates textual locality descriptions associated with biodiversity collections into geographic coordinates. The GEOLocate algorithm begins by standardizing the input locality string into commonly understood terms \citep{noauthor_geolocate_nodate}. It extracts details such as distances, compass directions, and significant geographic identifiers, which are used in lookups across various datasets such as place names, river miles, legal land descriptions, and highway-water body crossings to calculate geographic coordinates. Adjustments are made based on the parsed data. Further refinement of coordinates is achieved by scanning for water body names in the locality string and adjusting to the nearest point on the identified water body, enhancing accuracy for aquatic collections. The final coordinates are ranked, displayed digitally, and subjected to human verification or correction \citep{noauthor_geolocate_nodate}. GEOLocate is primarily developed for georeferencing data from the USA, Canada, and Mexico.

Both GEOLocate and BioGeomancer were developed initially in the early 2000s. GeoPick \citep{marcer_geopick_2023} is the latest addition to these georeferencing tools, developed by implementing the guidelines provided in \citet{chapman_georeferencing_2020}. It offers a comprehensive guide for georeferencing biological collections. A common feature among these tools—GEOLocate, BioGeomancer, and GeoPick—is their implementation of the point-radius method. Introduced by \citet{wieczorek_point-radius_2004}, the point-radius method does more than describing a location using a single coordinate pair; it indicates an area of uncertainty using a circle with a well-defined radius \citep{doherty_georeferencing_2011}. However, unlike its predecessors, GeoPick is not an automatic georeferencing tool; instead, it aids georeferencers in following standards and guidelines more effectively and in a user-friendly manner. 

\subsection{Large Language Models (LLMs)}

The emergence of LLMs has brought significant advancements to the field of Natural Language Processing (NLP). LLMs, primarily based on the decoder Transformer architecture, have evolved from the development of the GPT (Generative Pre-trained Transformer) models. As a result of the widespread success of these models, encoder-based approaches, such as BERT and XLNET, as well as encoder-decoder-based models such as BART and T5, have become less prominent. Characterised by their large number of parameters and immense training datasets, LLMs have shown high proficiency in comprehending and generating human language \citep{karanikolas2023large}. Fine-tuning pre-trained LLMs via supervised learning has become the key to achieving high performance in various tasks \citep{singhal2023large, ouyang2022training, zhang_enhancing_2023, ranjit2023retrieval}. Fine-tuning is the process of adapting a pre-trained LLM to a specific task or domain by continuing its training on a smaller, task-specific dataset. This approach leverages the model’s existing knowledge while allowing it to specialise in a particular application \citep{ziegler2019fine}. By adjusting the model’s parameters based on task-specific labelled data, fine-tuning enhances performance and enables LLMs to generate more accurate and contextually relevant outputs \citep{wei2021finetuned}. While GPT models \citep{openai2023gpt} are restricted to online inference and paid APIs, the state-of-the-art open source models such as Mistral \citep{jiang2023mistral}, Llama 1-4 \citep{touvron2023llama1, touvron2023llama2}, GLM \citep{zeng2022glm}, and Gemma \citep{team2024gemma} are often fine-tuned for downstream tasks, with accuracies comparable to those of the GPT models. 

However, due to the computational costs and complexity of fine-tuning very large parameter spaces, deploying and adapting LLMs to specific tasks via fine-tuning is challenging \citep{fujiwara_modify_2024}. To address this issue, several Parameter Efficient Fine-Tuning (PEFT) methods have been introduced \citep{hu2022lora, liu2021p, lester2021power, zaken2021bitfit}. Among those methods, Low-Rank Adaptation (LoRA) 
\citep{hu2022lora} has gained attention for its efficiency in fine-tuning by focusing on a limited set of parameters, thus lowering the total computational load. Taking one step further, QLoRA (Quantization and LoRA) presents an even more memory-efficient iteration of LoRA by quantizing the weights of the LoRA adapters to a lower precision \citep{dettmers2024qlora}. 

Prompt engineering is an essential skill for effective communication with LLMs \citep{white_prompt_2023}. A prompt is a set of instructions given to an LLM to tailor its behaviour and/or enhance its capabilities \citep{liu2023pre}. Empirical evidence shows that higher-quality prompts result in better performance across various tasks \citep{wei2022chain, yao2024tree}. Zero-shot prompting is the simplest prompting technique. It involves giving the model the task in the form of a prompt without any prior examples. Some of the prompting patterns discussed in the literature are the Persona pattern, Context Manager pattern, Thought Generation pattern, Template pattern, and Question Refinement pattern \citep{schulhoff2024prompt, white_prompt_2023}. 

The Persona pattern, which is a type of output customisation prompting, assigns an LLM a specific persona or role, guiding it to adopt a particular point of view when generating the output. This approach leverages the LLM’s ability to simulate expert-level understanding, thereby enhancing its performance in complex domains that require specialised knowledge \citep{white_prompt_2023, desmond2024exploring}. In contrast, the Context Manager pattern controls the contextual information, refining the response according to the given context.

Chain of Thought (CoT) prompting, which is a Thought Generation prompting pattern, has been identified as performing well in complex reasoning tasks \citep{wei2022chain}. This pattern guides the model through a step-by-step reasoning process to reach a conclusion. The simplest form of CoT, zero-shot CoT essentially involves adding the phrase `Let's think step by step" to the original prompt which has also shown better performance in reasoning tasks \citep{kojima2022large}. With each pattern having a different approach to define the task for the LLM, the best pattern depends on the task for which the LLM is used. 

In the context of georeferencing biological specimen data, most existing technologies have not kept pace with these advancements in NLP. Several studies demonstrate the use of early transformer models, notably BERT, for geolocating social media data as discussed earlier \citep{scherrer_heljuvardial_2020, simanjuntak2022we, lutsai2023geolocation}.  With modern LLMs becoming more powerful due to their decoder-based architecture, there is a motivation to apply them to the georeferencing task.

A few studies have utilized LLMs in related disciplines such as geospatial reasoning, geographic information systems, and geoscience, showing promising results \citep{mooney_rohin_2023, deng_k2_2023, roberts2023gpt4geo, yang_geolocator_2023, li_geolm_2023, zhang2023geogpt}. A recent study by \cite{hu2024toponym} has successfully employed LLMs for toponym resolution by fine-tuning the models to estimate the unambiguous references of toponyms. However, the authors acknowledge that LLMs alone are insufficient for comprehensive toponym resolution and incorporate additional geocoding services in their methodology to address inaccuracies in the LLM-generated results. Relatedly, \cite{hu2023geo} propose a geo-knowledge-guided GPT approach for disaster response that injects domain-specific geographic knowledge into prompts to extract complete location descriptions from Hurricane Harvey tweets. Their method substantially outperforms off-the-shelf NER baselines (reported as $>40\%$ improvement) and also improves markedly over default GPT models (reported as $\sim 76\%$ improvement), highlighting the practical value of explicit geographic guidance for strengthening LLM-based location understanding and extraction.

\cite{bhandari2023large}  evaluated whether LLMs are geospatially knowledgeable through three experimental approaches. They assessed the geospatial awareness of LLMs by analysing the use of geospatial prepositions in sentences. The study found that LLMs were able to correctly arrange cities that are physically closer to each other when prepositions such as `near' and `close to' were used to indicate proximity. Conversely, when the context involved the preposition `far', indicating a distant location, the generated cities tended to be farther apart. The study concludes that LLMs hold significant potential for supporting human tasks in geospatial reasoning and analysis, especially with targeted fine-tuning tailored to specific use cases. However, to the best of our knowledge, no existing studies have applied LLMs for georeferencing tasks to obtain geographic coordinates that best match the textual locality descriptions provided.

\section{Methodology}

\subsection{Data}

In this study, we utilise multiple biological collection datasets obtained from the Global Biodiversity Information Facility\footnote{http://www.gbif.org} (GBIF). GBIF is the largest online biodiversity data network in the world, containing biological collection records from many institutions worldwide. We selected multiple datasets from GBIF \citep{chr, mel, usa, mex}, which contain records from the regions of the USA, New Zealand, Australia, and Mexico. The USA, New Zealand, and Australia datasets are in English, while the dataset from Mexico is in Spanish. Since datasets available through GBIF follow the same data standards, these datasets share a common structure and can be preprocessed using the same steps. An occurrence record from GBIF contains a wealth of information, including details about the location where the samples were collected, information about the collector, and specifics of the collection event, such as temperature and altitude. Among this information, we extracted the columns `\emph{locality}', `\emph{decimalLatitude}', `\emph{decimalLongitude}', `\emph{countryCode}', and `\emph{stateProvince}', which are related to the location of the specimen found for our final dataset. `\emph{locality}' is the text that we are trying to georeference. `\emph{decimalLatitude}' and `\emph{decimalLongitude}' provide the original coordinates of the location described by `\emph{locality}', and we treat them as ground truth for georeferencing.  The country and state/province where the specimen was found were used as additional contextual information.

We extracted a subset of these datasets and applied several preprocessing steps: these included removing duplicate locality descriptions and eliminating records that do not have original coordinates. After preprocessing, we randomly selected approximately 30,000 records from each dataset for our initial experiments. Subsequently, we experimented with different dataset sizes. Since the datasets include only records georeferenced with the coordinates of the specimen collection site, these coordinates serve as the ground truth for model training. A study by \cite{yesson2007global} reports that 83\% of records in GBIF are annotated with accurate coordinates. The summary of the datasets used is provided in Table \ref{tab:sum_dataset}. Each country-specific dataset was divided into training (70\%), validation (15\%), and testing (15\%) sets for model training, validation, and testing, respectively.

\subsection{Prompt Engineering}
\label{sec:prompt-engineering}
Since LLMs have not previously been used in georeferencing text data with geographical coordinates, the most effective prompt structure for the task was identified by experimenting with multiple prompting patterns. During this experimental phase, we utilised ChatGPT-4 (OpenAI’s\footnote{https://openai.com/} GPT-4 model) accessed via the ChatGPT\footnote{https://chatgpt.com/} interface, chosen for its rapid inference capabilities and ease of access. We experimented with the following prompting patterns: Chain of Thought, Zero-Shot Chain of Thought, Context Manager, and Persona. These patterns were chosen for their applicability to the georeferencing context.  Table \ref{tab:prompts} offers a few examples of the different prompt patterns we tested on a single locality description, along with the corresponding errors in the predicted coordinates generated by ChatGPT-4. Later, in Section \ref{sec:GPT-analysis}, we compare the results of these prompting patterns with the latest GPT models.

ChatGPT-4 successfully returned a pair of coordinates in the majority of cases, yielding relatively acceptable results. Notably, this was the case even in Zero-Shot prompting, where the model was given the task without any prior examples or additional instructions (see Table \ref{tab:prompts}). ChatGPT-4 claimed to produce these coordinates by calculating offsets based on the approximate locations of the place names mentioned in the text (see Figure \ref{fig:chatgpt-response} in Appendix A). This suggests that LLMs already possess some understanding of the task at hand. However, when the LLM lacks knowledge of the geographical coordinates for place names mentioned in the description, it did not generate coordinates but instead provided follow-up instructions. 

Given the success of CoT prompting in various complex tasks involving reasoning, we constructed step-by-step instructions for the LLM to follow in solving the problem of georeferencing. This approach generated a coordinate pair to represent the geographical location described, adhering to the provided instructions. Nonetheless, developing a set of instructions that would enable the LLM to derive accurate results proved challenging, as manual georeferencing is typically a complex process that involves analysing maps, aerial imagery, and various spatial data layers to accurately pinpoint locations.

Following the Context Manager pattern, we conducted an experiment where we specified the country and state or province in which the specimen was found, thereby constraining the output to that specific region. This approach yielded reasonably good results with ChatGPT-4. We also explored the Persona pattern, where the LLM was instructed to assume the role of a georeferencer to derive coordinates from locality descriptions. Although this method aimed to leverage the LLM’s ability to mimic expert georeferencing skills, the results were less effective. The Persona pattern mostly led the LLM to explain the steps a georeferencer would take, rather than generating the coordinates independently.

\begin{figure}
    \centering
    \fbox{\includegraphics[width=0.8\linewidth]{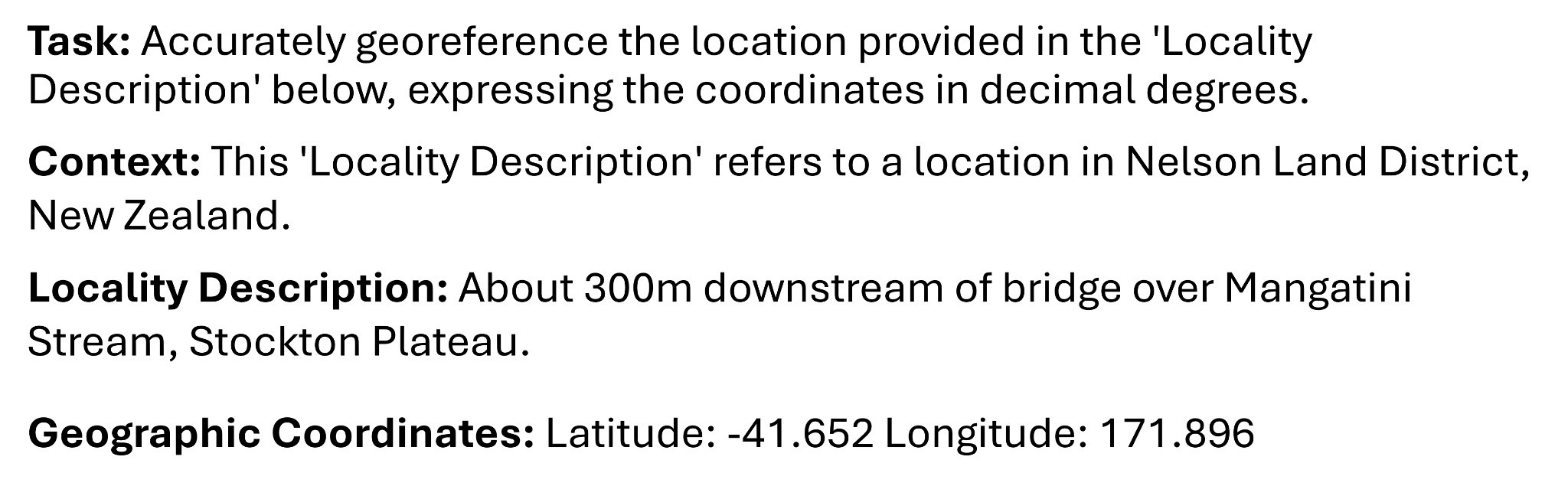}}
    \caption{Sample prompt used for fine-tuning the LLM. Both the training and validation datasets were structured using this pattern, with the original coordinates serving as the ground truth for supervised fine-tuning of the LLM. The test data was similarly formatted but without the inclusion of the coordinates.}
    \label{fig:sample-prompt}
\end{figure}

Through experimentation with various locality descriptions under different prompting patterns, we determined that the Context Manager pattern is the optimal prompt for our use case. This pattern is not only straightforward but also ensures that the produced coordinates accurately align with the specified geographic context. We then formatted the datasets according to the Context Manager pattern. In the training and validation prompts, the original coordinates are included as ground truth data, as illustrated in Figure ~\ref{fig:sample-prompt}. 

\begin{table}
\tbl{Sources of datasets extracted from GBIF for fine-tuning LLMs across different regions.}
{\begin{tabular}{ll} \toprule
Country & Source \\ \midrule
New Zealand & Manaaki Whenua - Landcare Research \\
USA & New York Botanical Garden Herbarium \\
Australia & Australasian Virtual Herbarium (AVH) \\
Mexico &  National Biological Collections of the Institute of Biology, UNAM \\ 
 
 \bottomrule
\end{tabular}}
\label{tab:sum_dataset}
\end{table} 

\begin{table}
\tbl{Overview of the experiments with different prompt patterns using ChatGPT-4.}
{\begin{tabular}{lll} \toprule
 Prompt Pattern & Prompt & Error (km) \\ \midrule
 Zero-shot & \begin{tabular}[c]{@{}l@{}}Accurately georeference the location provided in the \\`Locality Description' below, expressing the coordinates in decimal degrees.\\ Locality Description: 21 km west of Opotiki, south of Wainui Road\end{tabular} & 3.27km \\ \midrule
 \begin{tabular}[c]{@{}l@{}}Zero-shot\\ CoT\end{tabular} & \begin{tabular}[c]{@{}l@{}}Accurately georeference the location provided in the \\`Locality Description' below, expressing the coordinates in decimal degrees.\\ Locality Description: 21 km west of Opotiki, south of Wainui Road. \\ \textbf{Think step-by-step.}\end{tabular} & \begin{tabular}[c]{@{}l@{}}No coordinates generated. \\ Step by step guide to \\georeference is returned\end{tabular} \\ \midrule
 \begin{tabular}[c]{@{}l@{}}Chain of Thought\end{tabular} & \begin{tabular}[c]{@{}l@{}}Task: Convert the Locality Description below into decimal degree coordinates\\ Locality Description: 21 km west of Opotiki, south of Wainui Road.\\ Steps to follow: Start by identifying the decimal coordinates of the key reference point. \\Then, convert these coordinates into eastings and northings for precise adjustment. \\ Afterward, apply the specified distance and \\direction adjustments as per the locality description. \\ Finally, convert the adjusted eastings and northings back into decimal degree \\coordinates for the final location.\end{tabular} & 3.29km \\ \midrule
 Context Control & \begin{tabular}[c]{@{}l@{}}Accurately georeference the location provided in the \\`Locality Description' below, expressing the coordinates in decimal degrees.\\Context: This `Locality Description' refers to a location in North Island, New Zealand. \\ Locality Description: 21 km west of Opotiki, south of Wainui Road\end{tabular} & 3.03km \\ \midrule
 Persona Prompting & \begin{tabular}[c]{@{}l@{}}Act as a georeferencer and accurately georeference the location provided\\ in the `Locality Description' below, expressing the coordinates in decimal degrees. \\ Locality Description: 21 km west of Opotiki, south of Wainui Road\end{tabular} &  \begin{tabular}[c]{@{}l@{}}No coordinates generated. \\ Step by step guide to \\georeference is returned\end{tabular} \\ \bottomrule
\end{tabular}}
\label{tab:prompts}
\end{table}

\subsection{Fine-tuning LLM}

\begin{figure}
    \centering
    \fbox{\includegraphics[width=1\linewidth]{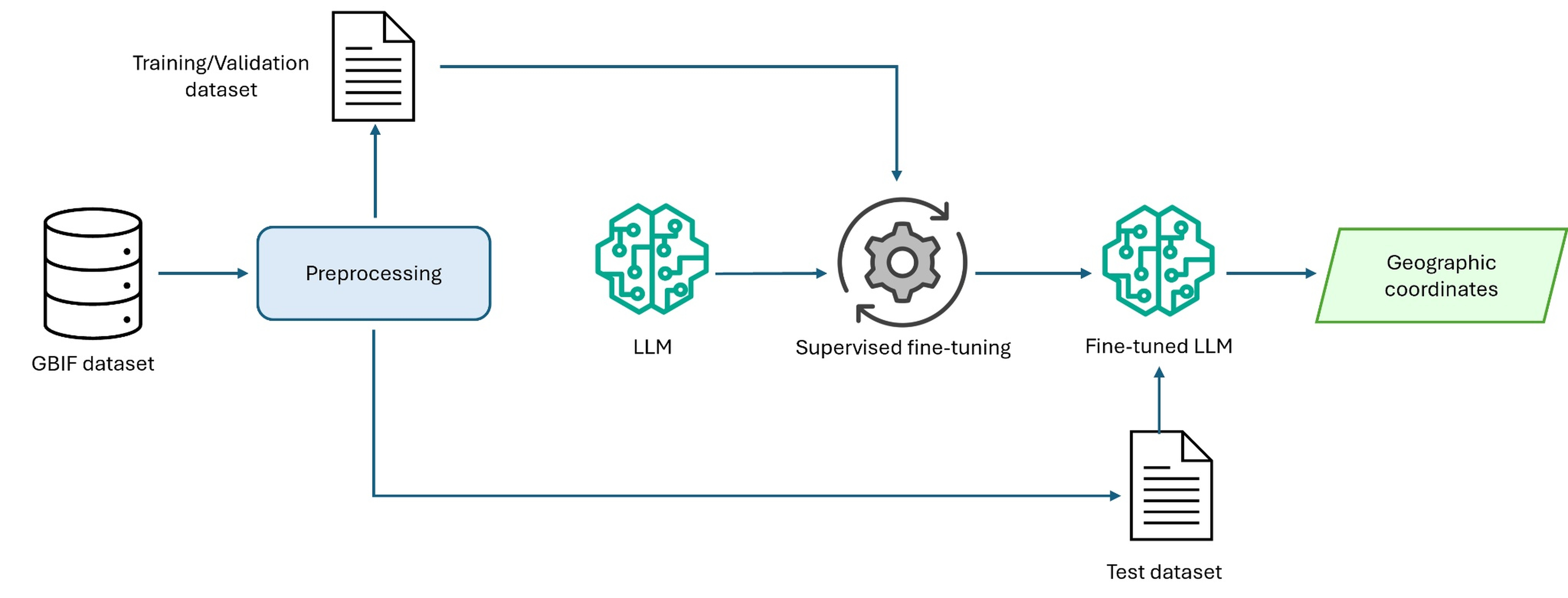}}
    \caption{Framework of fine-tuning an LLM for georeferencing biological specimen collection records.}
    \label{fig:methodology}
\end{figure}

Fine-tuning an LLM typically includes multiple steps. The process begins with preparing the dataset, followed by selecting and loading a pre-trained model. Before training, the text data must undergo tokenization, where it is broken into smaller units called tokens (which can be words, subwords, or characters) and then converted into numerical representations that the model can process. Tokenized data is used to train the LLM, during which the LLM’s parameters are adjusted, performance is monitored, and iterative refinements are made to improve accuracy. The steps we followed in this process are summarized in Figure~\ref{fig:methodology}.

To select the most suitable LLM for our task, we conducted initial experiments using several open-source pre-trained LLMs. 4-bit quantization and Low-Rank Adaptation (QLoRA) were applied to the base LLM to efficiently manage the large size and complexity of the LLM while maintaining high performance. This approach enabled us to fine-tune the model using a single NVIDIA A100 24 GB GPU, eliminating the need for large-scale computing infrastructure and making the solution more accessible and practical for scaling georeferencing tasks.

\subsection{Experiment Setup}

Initial experiments were conducted on the base version of the following open-source pre-trained LLMs: Llama2-7B, Llama3-7B, Gemma-7B, and Mistral-7B to identify the model that best matches the use case (we also tested instruct models, but they did not perform as well as the base models for this task). We chose open-source LLMs due to their accessibility, flexibility for customisation, and cost-efficiency compared to proprietary LLMs such as ChatGPT. These experiments identified the Mistral-7B base model as the most suitable for the task.

The key hyperparameters for model training were carefully selected to optimize performance while preventing overfitting, based on experiments with multiple settings. The learning rate was set to 2e-4, which offers a balance between convergence stability and training speed, mitigating divergence or oscillatory behaviour. The batch size was set to 32, a choice that offered stable gradient estimates without exceeding memory limits. For efficient fine-tuning, LoRA parameters were configured with a LoRA rank of 32 and LoRA alpha of 64. The rank specifies the dimensionality of the low-rank adaptation matrices, controlling their expressive capacity, while alpha rescales the output of the low-rank matrices before combining them with the frozen base weights, effectively controlling the contribution of the LoRA adaptation \citep{hu2022lora}. The training process consisted of three epochs. The low-rank matrices were iteratively updated at each batch step within each epoch using gradients calculated from the loss function. 

Checkpoints were established at regular intervals throughout the training process to facilitate continuous monitoring and potential retrieval of intermediate models. This fine-tuning process was performed separately for each region-specific dataset, resulting in a dedicated LLM fine-tuned for georeferencing within each region. The final models were saved and subsequently used for inference on the test datasets. Performance metrics were calculated to evaluate each model's accuracy in georeferencing locality descriptions.

\section{Evaluation}
\subsection{Baseline Systems}
Given the lack of publicly accessible and functional methods for georeferencing biological collection data, our selection of benchmarks was limited. While \cite{van_erp_georeferencing_2015} offer the most recent study on georeferencing biological specimen data, we were unable to use their methodology as a baseline because their application, `GeoImp,' is no longer available. Furthermore, the latest tweet georeferencing methodologies are also inapplicable to the domain of biological specimen data georeferencing, as they rely on metadata and social media user network data, in addition to the tweet text.

As a result, we compared the performance of our models against two georeferencing baselines: GEOLocate and a gazetteer-matching algorithm. GEOLocate is the only currently available tool for georeferencing locality descriptions of biodiversity collection records. For a given description, GEOLocate can return multiple coordinate pairs as potential geolocations, and these coordinates are ranked according to their probability of being correct \citep{noauthor_geolocate_nodate}. For our analysis, we selected the coordinate pair ranked as the most likely. 

Since many methods for georeferencing textual data primarily focus on identifying and geocoding place names through gazetteer-matching approaches, we implemented a gazetteer-matching algorithm as our second baseline (see Appendix B). This algorithm can be regarded as analogous to the spatial minimality disambiguation method of \cite{leidner2003grounding}. It involves the steps of place name recognition and resolution. To improve the accuracy of place name identification, we fine-tuned the spaCy\footnote{https://spacy.io/} model using 50 annotated locality descriptions of our data. The identified place names were then geocoded using a gazetteer-matching process. We experimented with both the GeoNames\footnote{https://www.geonames.org/} and Nominatim\footnote{https://wiki.openstreetmap.org/wiki/Nominatim} gazetteers to determine the most suitable option. During geocoding, we specified the country and state or province to initially disambiguate the place names. Given that a single locality description can contain multiple place names, it was necessary to disambiguate these names and identify the coordinates that best represent the overall locality description. To achieve this, we conceptually plotted all possible locations of the identified place names using the DBSCAN algorithm \citep{ester1996density} and identified the cluster with the highest concentration of points, assuming this cluster contained most of the place names mentioned in the description. To determine the final coordinate pair, we calculated the mean latitude and mean longitude of the points within the identified cluster.

In addition to these baselines, we also evaluated our fine-tuned LLM against three prominent GPT models, GPT-5, GPT-4.1 and GPT-4o, to assess its relative performance in the georeferencing task.

\subsection{Results and Discussion}
The mean and median of the SAE (Simple Accuracy Error) and the percentages of predicted locations that are within a 10km and 1km radius of the actual locations were used to evaluate the performance. The error was calculated by measuring the distance between the predicted location and the original location, utilising the Haversine formula. Table \ref{tab:comparison-with-geolocate} shows a comparison of the performance metrics between the fine-tuned Mistral model and baselines for each dataset.

\subsubsection{Performance Across Different Regions}

The fine-tuned LLM for the New Zealand region outperformed all baselines, correctly predicting 70.43\% of localities within a 10 km radius, 25.36\% within a 1 km radius and 7.71\% within 100 m. It also achieved a median SAE of 3.55 km and a mean SAE of 41.95 km. In comparison, GEOLocate reached only 45.39\% within 10 km and 10.45\% within 1 km, with substantially higher error values, while the gazetteer method performed worst overall.

For the Australian dataset, the fine-tuned LLM also surpassed the baselines, predicting 53.82\% of records within 10 km and 9.8\% within 1 km, with a median SAE of 8.52 km and a mean SAE of 51.28 km. While the performance did not reach the level achieved in New Zealand, the results for Australia were still clearly positive. In contrast, for the USA dataset, GEOLocate outperformed our LLM. GEOLocate predicted 26.82\% of localities within a 1 km radius, whereas our LLM predicted only 7.40\% within the same radius. The median and mean SAE given by GEOLocate also outperformed our model. The higher performance for the USA with GEOLocate might relate to the fact that GEOLocate is described as intended for the USA, Canada and Mexico regions \citep{noauthor_geolocate_nodate}. However, as demonstrated subsequently, our approach was found to be superior when applied to data for Mexico and for two states of the USA. 

Based on these results, we hypothesize that the lower performance in the USA dataset and the slightly reduced performance in the Australian dataset are due to the lower density of records relative to their large geographic extents. With fewer records per unit area, the training data may not provide sufficient coverage for the model to effectively learn relevant place names, recurring locality descriptions, and spatial language. If the availability of these features is a function of geographic area, then for a fixed number of training records, larger regions will yield fewer such signals per unit area, thereby reducing model effectiveness. To validate our hypothesis regarding the influence of record density and geographic extent, we designed two complementary experiments.

In the first experiment, we expanded the USA dataset to approximately 150,000 records and fine-tuned the LLM on 70\% of these data. The LLM demonstrated a significant improvement over the smaller dataset, accurately georeferencing 61\% of localities within a 10 km radius and 17.87\% within a 1 km radius. It achieved a median SAE of 5.91 km and a mean SAE of 36.48 km, while GEOLocate's performance remained consistent with that of the smaller dataset. These results indicate that increasing dataset size, and thereby effective record density, enhances the model’s effectiveness in interpreting locality descriptions and resolving place names and associated spatial semantics.

In the second experiment, we restricted training to smaller geographic regions using state-specific data. From the USA dataset, we fine-tuned separate LLMs with 20,962 records from New York and 12,441 from California. As shown in Table \ref{tab:comparison-with-geolocate}, the New York model predicted 84.86\% of coordinates within 10 km and 66.71\% within 1 km, while the California model achieved 66.26\% within 10 km and 28.7\% within 1 km, both outperforming the baselines. A similar pattern was observed for Australia, where a Victoria-specific model (29,568 records) predicted 81\% of localities within 10 km and 22.79\% within 1 km.

Together, these experiments support our hypothesis: the methodology is more effective in smaller regions with higher record density, and for larger regions, achieving comparable accuracy requires substantially more data.

Building on these region-specific results, we next examined whether the model's effectiveness extends across languages, since LLMs are trained on multilingual data \citep{qin2025survey, thellmann2024towards}. Given the global need for georeferencing biological specimen collection data, we were particularly interested in testing the linguistic generalizability of the model. Thus, we extended our experiments to test the model with Spanish-language data from the Mexico region. Our model, fine-tuned for the Mexico region with Spanish-language data, returned excellent results, predicting 75.82\% of coordinates within 10 km from the actual location and 51.41\% of localities within 1 km from the actual location. This model also achieved low error distances, with a median SAE of 0.89 km and a mean SAE of 20 km. GEOLocate tool, which is said to have been optimised for the Mexico region, only predicted 29\% records within a 10 km radius. These results suggest that our methodology effectively leverages the multilingual inference capability of LLMs, making it generalizable both regionally and linguistically.

\begin{table}[]
\tbl{Comparison of the results from the fine-tuned large language models with baseline systems across regional datasets. All datasets are in English, except for Mexico, which is in Spanish. The number of records shown refers to the entire dataset, which was subsequently divided into 70\% for training, 15\% for validation, and 15\% for testing.}
{\begin{tabular}{@{}lllllll@{}}
\toprule
 Dataset & No of records & Model & Accuracy@10km & Accuracy@1km & Med SAE & Mean SAE \\ \midrule
\multirow{3}{*}{New Zealand}  & \multirow{3}{*}{29,024} & GEOLocate & 45.39\% & 10.45\% & 12.24km & 129.62km \\ \cmidrule(l){3-7} 
                             &  & Gazetteer-based & 23.24\% & 5.30\% & 35.96km & 117.91km \\ \cmidrule(l){3-7}
                             & & Fine-tuned Mistral 7B & \textbf{70.43\%} & \textbf{25.36\%} & \textbf{3.55km} & \textbf{41.95km} \\ \midrule 
\multirow{3}{*}{USA}  & \multirow{3}{*}{29,566}       & GEOLocate & \textbf{56.40\%} & \textbf{26.82\%} & \textbf{5km} & \textbf{62.38km} \\ \cmidrule(l){3-7} 
                             & & Gazetteer-based & 12.81\% & 3.47\% & 584km & 1062km \\ \cmidrule(l){3-7}
                             & & Fine-tuned Mistral 7B & 35.24\% & 7.40\% & 19.18km & 63.9km \\ \midrule
\multirow{3}{*}{Australia} & \multirow{3}{*}{29,566}  & GEOLocate & 41.89\% & 6.85\% & 16.28km & 237km \\ \cmidrule(l){3-7} 
                             & & Gazetteer-based & 16.16\% & 1.91\% & 18.25km & 345.92km \\ \cmidrule(l){3-7}
                             & & Fine-tuned Mistral 7B & \textbf{53.82\%} & \textbf{9.80\%} & \textbf{8.52km} & \textbf{51.28km} \\ \midrule 
\multirow{3}{*}{New York (USA)} & \multirow{3}{*}{20,962}   & GEOLocate & 55.73\% & 32\% & 3.52km & 79km \\ \cmidrule(l){3-7} 
                             & & Gazetteer-based & 31.98\% & 9.63\% & 11.82km & 488km \\ \cmidrule(l){3-7}
                             & & Fine-tuned Mistral 7B & \textbf{84.89\%} & \textbf{66.71\%} & \textbf{0.083km} & \textbf{17.59km} \\ \midrule 
\multirow{3}{*}{California (USA)} & \multirow{3}{*}{12,441} & GEOLocate & 50.78\% & 20.78\% & 8.94km & 88km \\ \cmidrule(l){3-7} 
                             & & Gazetteer-based & 15.48\% & 2.09\% & 24.17km & 158.24km \\ \cmidrule(l){3-7}
                             & & Fine-tuned Mistral 7B & \textbf{66.26\%} & \textbf{28.70\%} & \textbf{4.17km} & \textbf{40.20km} \\ \midrule 
\multirow{3}{*}{Victoria (Australia)} & \multirow{3}{*}{29,568}   & GEOLocate & 48.82\% & 8.22\% & 10.73km & 169.44km \\ \cmidrule(l){3-7} 
                             & & Gazetteer-based & 20.26\% & 3.09\% & 12.76lm & 264.11km \\ \cmidrule(l){3-7}
                             & & Fine-tuned Mistral 7B & \textbf{81.76\%} & \textbf{22.79\%} & \textbf{2.94km} & \textbf{11.97km} \\ \midrule 
\multirow{3}{*}{Mexico}  & \multirow{3}{*}{32,116}    & GEOLocate & 29.85\% & 10.43\% & 54km & 251km \\ \cmidrule(l){3-7} 
                             & & Gazetteer-based & 15.94\% & 2.71\% & 82.66km & 313.11km \\ \cmidrule(l){3-7}
                             & & Fine-tuned Mistral 7B & \textbf{75.82\%} & \textbf{51.41\%} & \textbf{0.89km} & \textbf{20km} \\ \bottomrule 
\end{tabular}}
\label{tab:comparison-with-geolocate}
\end{table}

\subsubsection{Performance Compared to GPT Models}
\label{sec:GPT-analysis}

In addition to comparing our fine-tuned LLMs with existing georeferencing baselines, we conducted experiments using the latest and most prominent GPT models (GPT-5 \citep{openai_gpt5_2025}, GPT-4.1 \citep{openai_gpt41_2025} and GPT-4o \citep{openai_gpt4o_2024}) with the prompting strategies shown in Table~\ref{tab:prompts}. These experiments were performed via the OpenAI API\footnote{https://platform.openai.com/}. Due to the proprietary nature and high cost associated with these services, we evaluated only a random subset of 500 samples from the New Zealand test dataset, rather than the full test datasets for each region. The corresponding results are presented in Table~\ref{tab:gpt-comparison}. We have also included the results from the Mistral-7B model without fine-tuning, as well as our fine-tuned Mistral-7B model.

Our fine-tuned LLM significantly outperformed all GPT variants, including the newest GPT-5 model. This suggests that the proprietary models, regardless of how intelligent they have become or how much their reasoning capabilities have improved, are currently limited in their ability to perform the type of spatial reasoning required for accurate georeferencing. The substantial performance margin highlights the effectiveness of our domain-specific fine-tuning approach. However, it is worth noting that GPT models with the context control prompt achieved the best mean SAE values overall, with the lowest recorded by GPT-4o. 

Moreover, these experiments highlight the importance of prompt design. As shown in Table~\ref{tab:gpt-comparison}, the context control prompting pattern consistently yielded the best performance for both GPT-4.1 and GPT-4o. However, with the newest GPT-5 model, the Zero-shot CoT prompts produced slightly better results, likely due to the enhanced reasoning capabilities (\textit{GPT-5 thinking}) introduced. Overall, these findings validate our prompting strategy. 

\begin{table}[]
\tbl{Comparison of results from the fine-tuned Mistral-7B model and GPT models on the New Zealand test set.}{
\begin{tabular}{@{}llllll@{}}
\toprule
LLM & Prompt Pattern & Accuracy@10km & Accuracy@1km & Med SAE & Mean SAE \\ \midrule
 \multirow{5}{*}{GPT-4o} 
 & Zero-shot       & 52\% & 6.8\% & 9.12km   & 476km    \\ \cmidrule(l){2-6}
 & Zero-shot CoT   & 48.4\%    &  6.8\%   &   10.26km  &  373.76km        \\ \cmidrule(l){2-6}
 & CoT             & 48\% & 6.2\% & 10.69km   & 501.76km    \\ \cmidrule(l){2-6}
 & Context Control & 54.8\% & 8.4\% & 8.69km  & \textbf{45.33km}  \\ \cmidrule(l){2-6}
 & Persona         & 48.2\% & 6.4\% & 10.59km   & 452.79km \\ \midrule
 \multirow{5}{*}{GPT-4.1} 
 & Zero-shot       & 45.4\% & 6.2\% & 11.72km  & 399.14km \\ \cmidrule(l){2-6}
 & Zero-shot CoT   & 45\% & 6.4\% & 11.42km     & 340.63km \\ \cmidrule(l){2-6}
 & CoT & 48.6\% & 6.4\% & 11.42km & 340.63km \\ \cmidrule(l){2-6}
 & Context Control & 50.8\% & 7\% & 9.9km  & 51.18km  \\ \cmidrule(l){2-6}
 & Persona         & 47.6\% & 7.8\% & 10.92km     & 405.33km \\ \midrule
 \multirow{5}{*}{GPT-5} 
 & Zero-shot       & 43.57\% & 6.2\% & 13.26km  & 333km \\ \cmidrule(l){2-6}
 & Zero-shot CoT   & 59.4\% & 10\% & 7.17km     & 159.13km \\ \cmidrule(l){2-6}
 & CoT & 57.6\% & 9.6\% & 7.27km & 139.59km \\ \cmidrule(l){2-6}
 & Context Control & 57.4\% & 8.8\% & 7.81km  & 50.94km  \\ \cmidrule(l){2-6}
 & Persona         & 36.27\% & 5.4\% & 18.41km     & 416.76km \\ \midrule
  \multirow{1}{*}{Mistral 7B} 
 & Context Control & 5.23\% & 0.48\% & 49.78km  & 1120km \\ \midrule
 \multirow{1}{*}{Fine-tuned Mistral 7B} 
 & Context Control & \textbf{71\%} & \textbf{31\%} & \textbf{3.81km} & 52.01km \\ \bottomrule
\end{tabular}}
\label{tab:gpt-comparison}
\end{table}

\subsubsection{Effect of Locality Description Length on Performance}

\begin{table}[]
\tbl{Examples of locality descriptions with different lengths (in characters).}
{\begin{tabular}{@{}ll@{}}
\toprule
\textbf{Length} & \textbf{Locality Description} \\ \midrule                 
\multirow{3}{*}{\textless 30}     & Off Great Barrier Island \\ \cmidrule(l){2-2}
                                  & near Gulf Harbour \\ \cmidrule(l){2-2}
                                  & Beach near Devonport    \\ \midrule
\multirow{3}{*}{30-60}            & Three Kings Group, West Island. Off W side of S summit    \\    \cmidrule(l){2-2}                                                                   & Stream at Te Moori, 3 miles south of Kaeo, Northland     \\  \cmidrule(l){2-2}                                                                     & Jacko's pond, 3 km south of Coromandel                                    \\ \midrule
\multirow{3}{*}{60-90}            & Little Barrier Island, Awaroa Stream, about 150m from the sea   \\  \cmidrule(l){2-2}
                                  & Gully facing sea, on bluffs opposite cottage, 2km southwest of Kaihoka Lakes, Nelson    \\  \cmidrule(l){2-2}           
                                  & Milford, near beach at end of Milford Road, Waitemata County, North Island      \\ \midrule
\multirow{3}{*}{90-120}           & \begin{tabular}[c]{@{}l@{}}South Island, Fiordland National Park, W of Homer Tunnel, 9 mi SE of Milford Sound on Hw. 94\end{tabular}      \\  \cmidrule(l){2-2}
                                  & \begin{tabular}[c]{@{}l@{}}Central Volcanic Plateau: c. 2 km S of National Park, forest margin near grazed meadow 300m W of Mountain\\ Hights Lodge.\end{tabular}                       \\ \cmidrule(l){2-2}
                                  & South Island, Westland, 0.6 miles W. of Rahu Saddle on Rte. 7, Terrestrial on roadside bank                             \\ \midrule
\multirow{3}{*}{\textgreater 120} & \begin{tabular}[c]{@{}l@{}}Knightia Railway Reserve and adjoining grazed bush containing abundant moss-covered small shrubs, \\ on Morrison Farm about 4 km by road NW of Mataroa, itself on Mataroa Rd NW of Taihape\end{tabular} \\ \cmidrule(l){2-2}
                                  & \begin{tabular}[c]{@{}l@{}}Nelson Lakes National Park, St Arnaud, picnic/camping area on lake shore at Kerr Bay, \\ about 100 m towards DoC Headquarters from Black Stream bridge\end{tabular}                                     \\ \cmidrule(l){2-2}
                                  & \begin{tabular}[c]{@{}l@{}}Farm trail north side of Parau Stream, 200 yards (182m) above bridge across tributary, W\\ aimana Valley, inland Bay of Plenty, North Island, New Zealand. \end{tabular}   \\ \bottomrule                    
\end{tabular}}
\label{tab:ex-locality}
\end{table}

Another aspect we explored in our research was the impact of the character length of a locality description in a collection record on LLM's performance. In our dataset, we observed that longer descriptions tend to be more complex, often detailing the location of a specimen in relation to multiple distances and directions, referencing various locations. Examples of locality descriptions categorised by length are presented in Table \ref{tab:ex-locality}. 

\begin{figure}
    \centering
    \fbox{\includegraphics[width=0.8\linewidth]{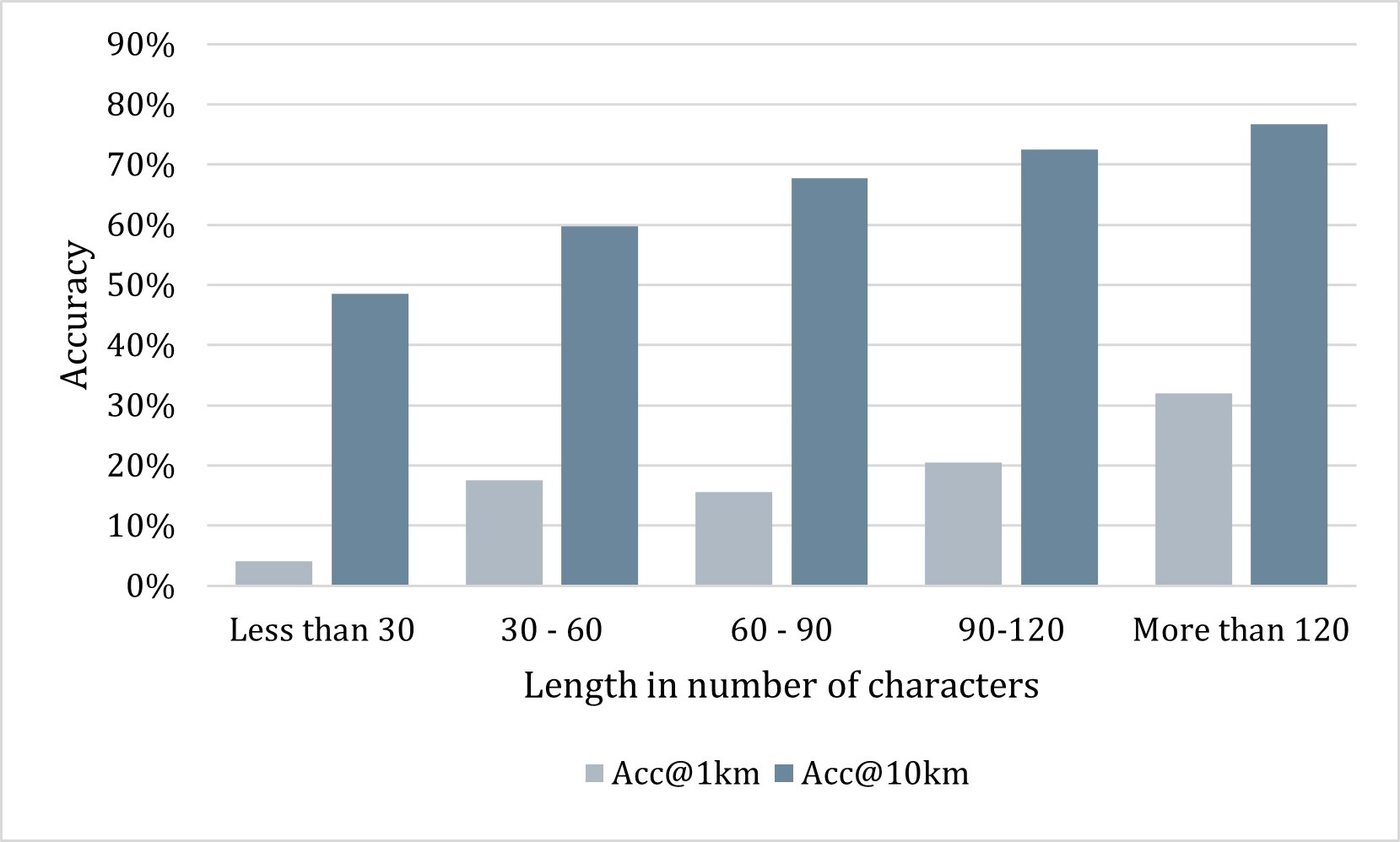}}
    \caption{Distribution of accuracy in error distance by the description length for the New Zealand dataset.}
    \label{fig:distance-accuracy}
\end{figure}

Table \ref{tab:length-comparison} presents a comparison of the results of our New Zealand fine-tuned LLM (reported previously in Table \ref{tab:comparison-with-geolocate}), grouped by description length. There is an incremental improvement in the accuracy of the prediction when the length of the description increases. The bar chart in Figure \ref{fig:distance-accuracy} further complements our observation. To investigate the statistical basis of this pattern, we analysed the correlation between error distance and description length using Spearman’s rank correlation. This method was chosen because it does not assume linearity or normally distributed variables and is robust to outliers. The analysis revealed a moderately negative correlation ($\rho = -0.19$, $p \ll 0.001$), indicating that longer descriptions are associated with reduced error, although the modest effect size suggests that description length is only one of several factors influencing prediction accuracy.

\begin{table}[]
\tbl{Comparison of prediction accuracy of the fine-tuned LLM by description length in the New Zealand test set.}
{\begin{tabular}{@{}lllll@{}}
\toprule
Length in number of characters & Acc@1km          & Acc@10km         & Median SAE      & Mean SAE         \\ \midrule
Less than 30            & 4.07\%           & 48.54\%          & 11.52km         & 55.85km          \\
30 - 60                 & 17.60\%          & 59.70\%          & 6km             & 66.45km          \\
60 - 90                 & 15.61\%          & 67.81\%          & 5.03km          & 26.37km          \\
90 - 120                & 20.55\%          & 72.59\%          & 3.78km          & \textbf{24.27km} \\
More than 120           & \textbf{32.01\%} & \textbf{76.77\%} & \textbf{2.40km} & 25.16km          \\ \bottomrule
\end{tabular}}
\label{tab:length-comparison}
\end{table}

\begin{figure}
    \centering
    \fbox{\includegraphics[width=1\linewidth]{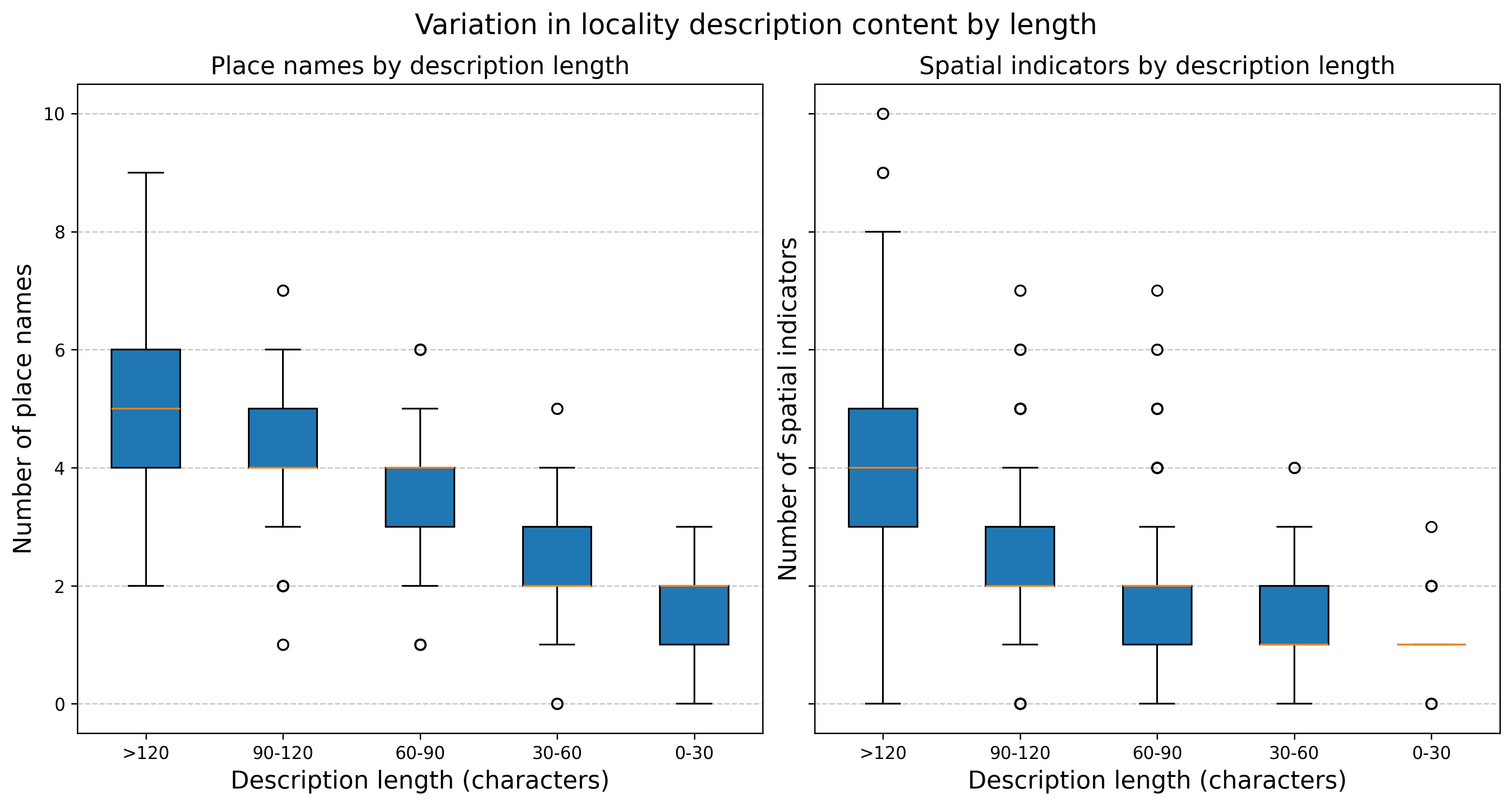}}
    \caption{Distribution of locality description features by length: (a) number of place names and (b) number of spatial indicators.}
    \label{fig:boxplot-si-pn}
\end{figure}

To clarify the source of these gains, we examined the description content by length. We looked into the place names and spatial indicators present in the locality descriptions. Spatial indicators are textual elements that express geographic relationships, providing cues to locate a place in reference to other geographic entities. For example, in the locality description \textit{`10 km north of Lake Wanaka, 1 km north of Makarora, near Pipson Creek’}, there are three place names (\textit{`Lake Wanaka'}, \textit{`Makarora'}, \textit{`Pipson Creek'}) that act as reference points, while the three spatial indicators (\textit{`10 km north of'}, \textit{`1 km north of'}, \textit{`near'}) specify the relative positions needed to infer the exact location. Longer localities contained more place names and spatial indicators as revealed by the box-plots in Figure \ref{fig:boxplot-si-pn}. Correlation analysis confirmed that the number of place names was moderately and negatively associated with error distance (Spearman $\rho = –0.16, p \ll 0.001$), indicating that descriptions with more toponyms tend to yield more accurate predictions.

\begin{figure}
    \centering
    \fbox{\includegraphics[width=1\linewidth]{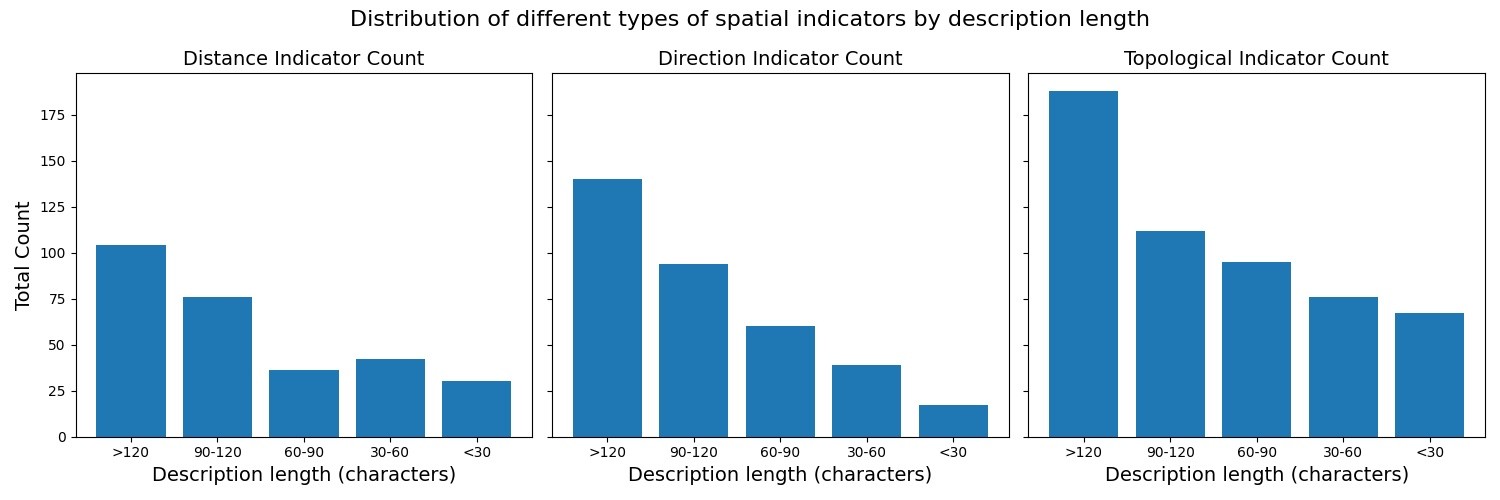}}
    \caption{Distribution of spatial indicators (distance, direction, and topological) as a function of locality description length.}
    \label{fig:si-count}
\end{figure}

We further disaggregated spatial indicators into three types: directional (e.g., \textit{`north of'}, \textit{`SE side of'}, \textit{`west-facing'}), distance (both quantitative, e.g., \textit{`5 km N of Wellington'}, and qualitative, e.g., \textit{`near the river'}), and topological (e.g., \textit{`on stone'}, \textit{`at the base'}). As shown by the box-plots in Figure~\ref{fig:si-count}, directional and distance cues increased sharply with description length, particularly beyond 90 characters, while topological indicators remained relatively frequent across all bins but peaked in the longest descriptions. These findings explain the observed performance gains: richer descriptions do not simply add verbosity but supply both a higher density of place names and a greater variety of spatial relations, which collectively improve LLM inference.

Together, these analyses demonstrate that description length is a meaningful predictor of georeferencing success, not because of length itself but because longer records tend to encode more place names and spatial relationships. Our results suggest that the fine-tuned LLM effectively leverages this additional context to reduce spatial error, underscoring the importance of detailed specimen metadata for accurate automated georeferencing. This interpretation is consistent with \cite{ji2025foundation}, who show that LLMs can interpret formal topological relations with moderate accuracy, but frequently confuse conceptually adjacent predicates, indicating that topological cues alone can remain ambiguous. In our setting, longer specimen records mitigate this ambiguity by providing more place anchors and stronger directional/distance constraints, enabling more precise coordinate inference.

Table \ref{tab:cal-comparison} lists the ten localities where GEOLocate exhibited the highest georeferencing errors in the California state dataset. These errors are measured as the distance between the predicted and actual locations. The table compares these errors with those produced by our fine-tuned LLM for the same localities. Notably, the localities that produced the poorest results with GEOLocate, are characterized by long descriptions containing multiple spatial terms and place names. In 9 out of these 10 instances, our model achieved significantly lower error margins. This finding further demonstrates that our method provides more accurate results in handling complex locality descriptions compared to current georeferencing approaches.

\begin{table}[]
\tbl{Georeferencing error comparison between GEOLocate and the fine-tuned Mistral 7B LLM for the ten worst-performing localities by GEOLocate in the California state dataset.}
{\begin{tabular}{@{}lll@{}}
\toprule
Locality Description & \multicolumn{1}{l}{\begin{tabular}[c]{@{}l@{}}Error with \\ GEOLocate \end{tabular}} & \multicolumn{1}{l}{\begin{tabular}[c]{@{}l@{}}Error with \\ Fine-tuned Mistral 7B \end{tabular}} \\ \midrule
\begin{tabular}[c]{@{}l@{}}Amargosa Desert, along dirt rd sw. of Calif.-Nev. state line from Calif. \\ Hwy 127 toward abandoned Tonopah \& Tidewater Railroad. \\ Drainage Basin: Amargosa; U. S. Atomic Energy Commission's Nevada \\ Test Site and vicinity.\end{tabular} & 1891.73km                                & 49.60km                                              \\ \midrule
Head of Etna Creek, Marble Mountains    & 997.35km  & 24.39km  \\ \midrule
Cleveland National Forest, Black Canyon Rd, 6km N of Ramona  
                                        & 986.59km  & 6.60km  \\ \midrule
Cleveland National Forest, Palomar district, Black Canyon Rd, \\7km S of junction with Rd 12S05                                                                                           & 982.91km  & 27.92km  \\ \midrule
Vicinity of Leonard Hot Spring, E side of Surprise Valley.                                                       & 971.70km  & 26.10km  \\ \midrule
\begin{tabular}[c]{@{}l@{}}Mohave Desert, northeast of the San Bernardino Mountains, \\ 9 airline miles north-northeast of junction of \\ Calif. Highway 62 and highway to Lucerne Valley (Yucca Valley town).\end{tabular}                     & 970.43km  & 915.58km \\ \midrule
Pyramid Point, forming a distinct zone in the middle zone \\of the litoral region, on rocks                                                                                                 & 948.77km  & 0.02km   \\ \midrule
Vail Lake area, lower slopes of Agua Tibia Mtn. along Hwy 79, \\E of USFWS Dripping Springs Guard Sta.                                                                                      & 937.70km  & 0.48km   \\ \midrule
Just S.W. of sand hills, 5 mi. N.E. of Glamis.                                                                   & 887.96km  & 5.78km   \\ \midrule
927 Candlelight Place, adjacent to N terminus of Cass St                                 & 842.72km  & 0.78km  \\ \bottomrule                                 
\end{tabular}}
\label{tab:cal-comparison}
\end{table}

\subsubsection{Effect of Prompt Wording Variations on Fine-tuning Performance}
As detailed in Section \ref{sec:prompt-engineering} and Table \ref{tab:prompts}, we initially experimented with several prompt patterns and identified the Context Control prompt pattern as the most appropriate for our use case. Accordingly, we used this pattern to structure the input data for training the fine-tuned LLMs. To explore whether the specific wording of prompts within this pattern influences the model performance of the fine-tuned LLM, we further extended our experiments to test different phrasings of the same Context Manager pattern. We used the New Zealand dataset (29,024 records) and compared the outcomes of the original prompt with two alternative phrasings. The results of these experiments are presented in Table \ref{tab:ctx-prompts}, where we observed minor differences in performance across the variations.

The first prompt represents the original wording used in our experiments. The second version introduces a slight modification to the instructional phrasing. The third prompt combines the Context Manager pattern with elements of the Persona pattern by explicitly defining the user’s persona alongside the context. Among these, the latter two variations showed a slight improvement in results. Accuracy within 10 km ranged only from 70.43\% to 71.72\%, and accuracy within 1 km varied between 25.36\% and 26.23\% across the three prompt versions. Median and mean errors likewise differed only marginally (3.38–3.55 km and 40–42.38 km, respectively). These results suggest that while prompt wording may have a minor influence on performance, it is not a dominant factor in the context of our fine-tuning setup.

\begin{table}[htbp]
\tbl{Comparison of results from variations of context manager pattern prompts for the New Zealand dataset.}{
\begin{tabular}{@{}p{8cm}cccc@{}} 
\toprule
\textbf{Sample Prompt} & \textbf{Acc@10km} & \textbf{Acc@1km} & \textbf{Median} & \textbf{Mean} \\ 
\midrule

Accurately georeference the location provided in the `Locality Description' below, expressing the coordinates in decimal degrees. \newline\textbf{Context}: This `Locality Description' refers to a location in Canterbury Land District, New Zealand. \newline\textbf{Locality Description}: `Inverary', Mt Somers, Mid Canterbury, above Blondin Stream. 
& 70.43\% & 25.36\% & 3.55 km & 41.95 km \\ 

\midrule

Determine the precise geographic coordinates for the location described in the `Locality Description' below, providing the latitude and longitude in decimal degrees. \newline\textbf{Context}: The described locality is situated in Canterbury Land District, New Zealand. \newline\textbf{Locality Description}: `Inverary', Mt Somers, Mid Canterbury, above Blondin Stream.
& \textbf{71.72\%} & 26.05\% & \textbf{3.38 km} & 42.38 km \\ 

\midrule

\textbf{You are a GIS specialist} verifying geographic data for locations in New Zealand. Your task is to determine the precise latitude and longitude (in decimal degrees) for the locality described below. \newline\textbf{Region}: Canterbury, New Zealand \newline\textbf{Locality Description}: `Inverary', Mt Somers, Mid Canterbury, above Blondin Stream.
& 71.12\% & \textbf{26.23\%} & 3.48 km & \textbf{40 km} \\ 

\bottomrule
\end{tabular}}
\label{tab:ctx-prompts}
\end{table}

\subsubsection{Evaluation of Model Sensitivity to Spatial Distance Information}

Many descriptions in our dataset include quantitative spatial distance values referenced to a place name, such as `\emph{200 metres west of Mangatoetoe Stream}' or `\emph{6 km SSE of Westport}'. To assess whether our fine-tuned LLM makes meaningful use of these numerical distance references in geolocation prediction, we conducted a targeted evaluation.

We randomly selected 100 records from the New Zealand test dataset and identified 59 descriptions containing explicit quantitative spatial information. These distance values were then manually removed from the text, with care taken to preserve the overall semantic meaning and spatial context. For example, the description ‘\textit{30 miles S of Auckland City}’ was revised to ‘\textit{S of Auckland City}’ by removing the quantitative distance ‘\textit{30 miles}’.

\begin{figure}[h]
    \centering
    \fbox{\includegraphics[width=0.8\linewidth]{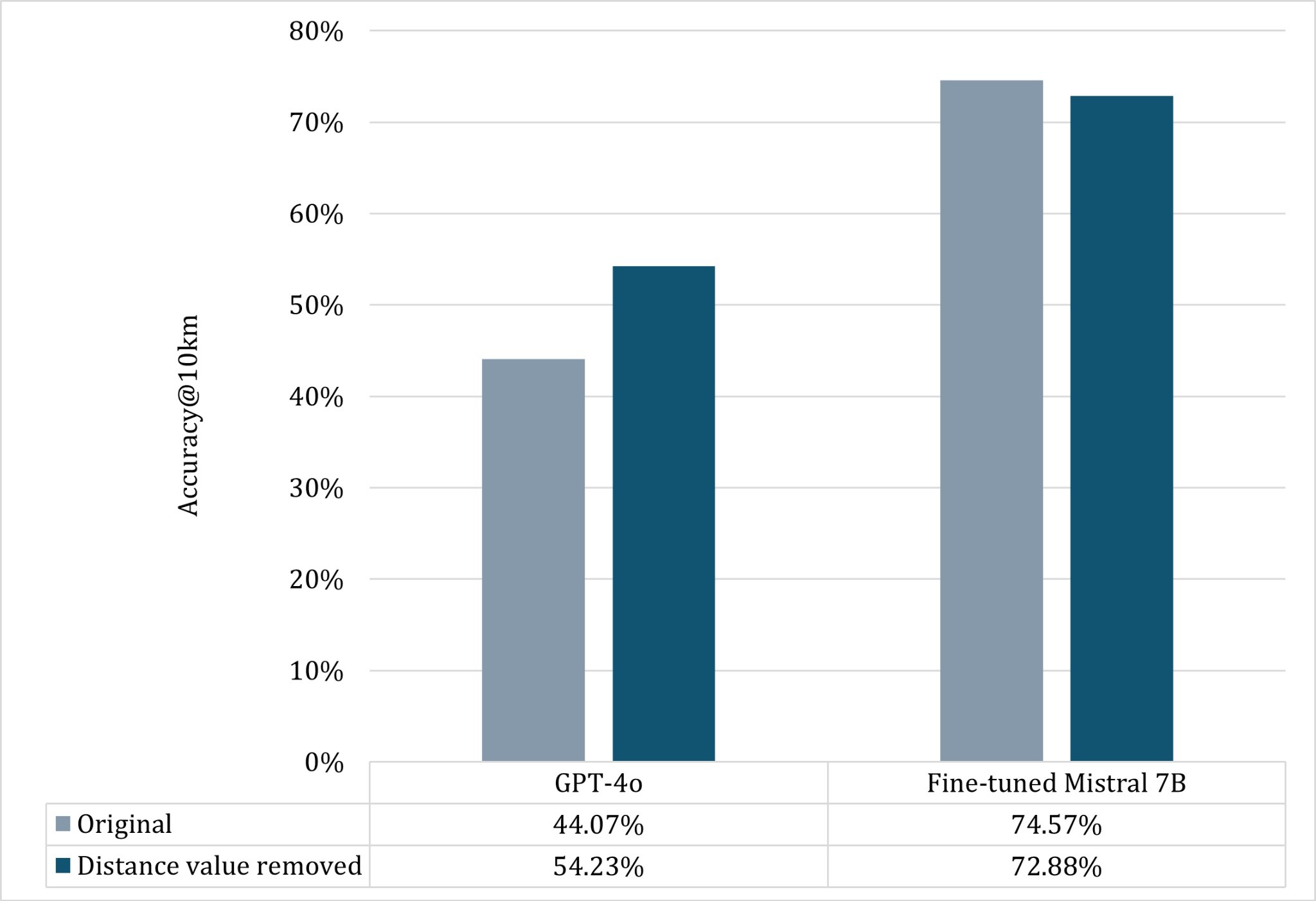}}
    \caption{Comparison of sensitivity to spatial distance information between GPT-4o and the fine-tuned Mistral 7B models in predicting coordinates, evaluated on a subset of the New Zealand test set.}
    \label{fig:distance-val-effect}
\end{figure}

We evaluated model performance on both the original and modified versions of these records using GPT-4o and our Mistral 7B model fine-tuned for the New Zealand region. Results are illustrated in the bar chart in Figure~\ref{fig:distance-val-effect}. Interestingly, GPT-4o exhibited unexpected behaviour, its prediction accuracy was higher when the distance values were removed. In contrast, the fine-tuned Mistral model behaved as expected, showing a slight decline in accuracy when distance expressions were removed. This suggests that the fine-tuned LLM is more attuned to leveraging such spatial cues during prediction. 

Despite the use of a smaller dataset, these results indicate that our fine-tuned LLM is capable of interpreting quantitative spatial cues, reinforcing its potential for accurate georeferencing of biodiversity collection records.

\subsubsection{Analysis of Transfer Learning Capabilities of LLMs}
\label{sec:transfer-learning}

To evaluate the transfer learning (TL) capabilities of our approach, we tested whether a model fine-tuned on data from one region could be applied to another, and whether combining or adapting datasets could improve performance.

We first tested whether a model fine-tuned exclusively on Australian data could be applied directly to New Zealand. The model was able to generate coordinates, but the predictions exhibited very high error margins (see Table \ref{tab:transfer-learning-nz-aus}). This indicates that while training on data from another country may help the LLM capture the general semantics of the task, it is insufficient for accurate coordinate prediction without region-specific information.

We next evaluated multi-region training configurations combining different proportions of Australian and New Zealand training data (see Table \ref{tab:transfer-learning-nz-aus}) and testing on the New Zealand test dataset. Adding 20\% New Zealand training data to an Australia-fine-tuned LLM reduced the mean error to 57.27 km, performing well above the baseline (GEOLocate). A balanced dataset (50\% AUS / 50\% NZ) lowered it further to 47.9 km mean error. Training with predominantly New Zealand data (20\% AUS / 100\% NZ) achieved the best result of 39 km, only slightly better than New Zealand–only training (41.95 km). These results suggest that modest amounts of local data can markedly improve the transferability, but cross-region data alone contributes little beyond what region-specific training achieves.

We also tested the generalisability of the New Zealand–fine-tuned LLM across different datasets within the same region. Specifically, we used the model originally fine-tuned on the New Zealand Allan Herbarium (CHR) \citep{chr} dataset to georeference a randomly selected sample of 5,000 records from both the New Zealand Arthropod Collection (NZAC) \citep{nz_nzac} and the New Zealand National Forestry Herbarium \citep{nz_forestry}. The model achieved mean SAEs of 23.2 km (NZAC) and 43.4 km (Forestry Herbarium), comparable to the performance on the original CHR dataset (41.95 km). Median errors were consistently low across datasets (3.55–7.99 km). These results suggest that the model can generalise reasonably well across independently curated datasets within a region, indicating a degree of transferability beyond the dataset it was originally fine-tuned on.

\begin{figure}[h]
    \centering
    \fbox{\includegraphics[width=0.8\linewidth]{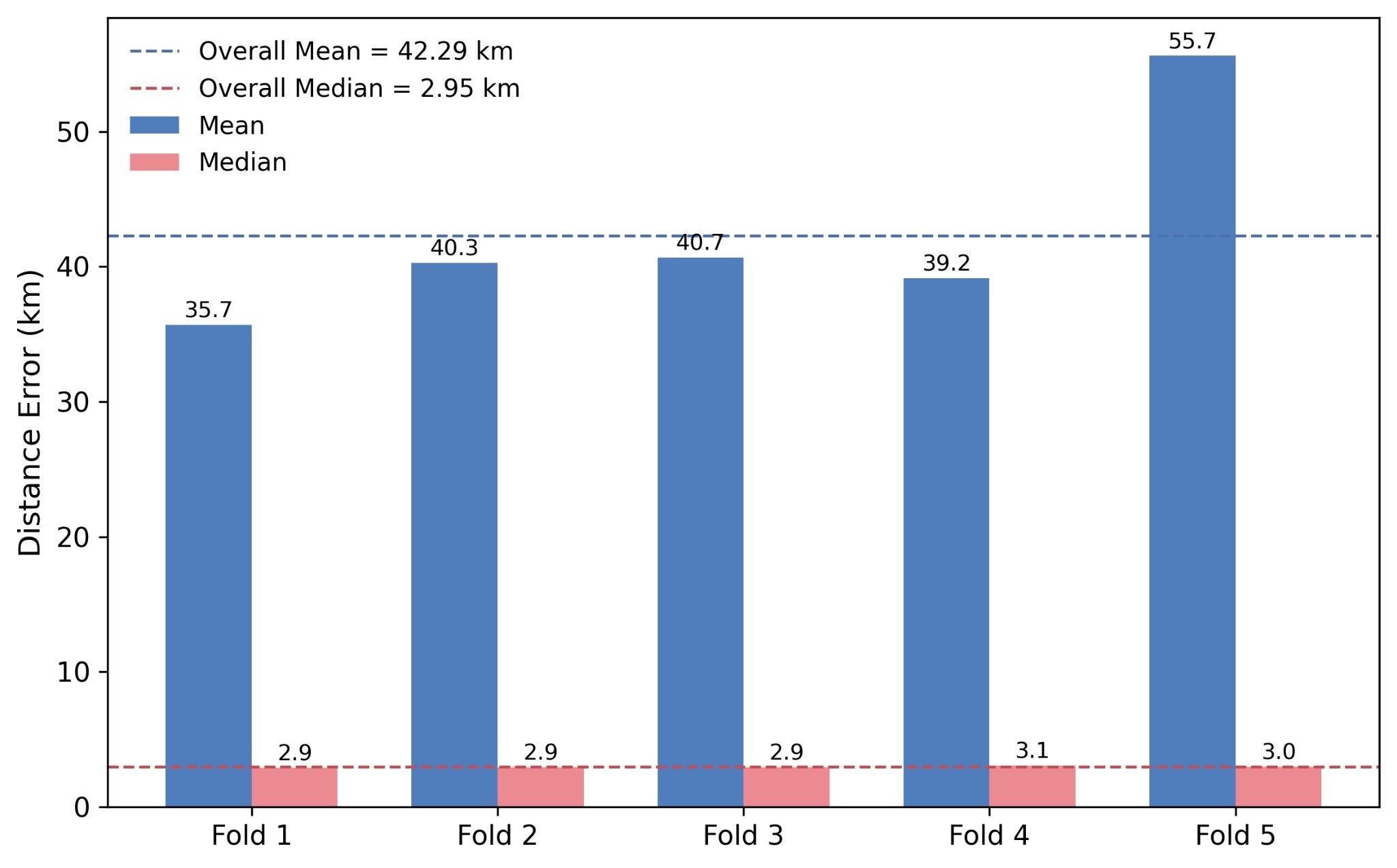}}
    \caption{Cross-validation results of the fine-tuned Mistral-7B model for New Zealand.}
    \label{fig:cv-nz-bar}
\end{figure}

To further validate this robustness, we conducted a 5-fold cross-validation experiment on the original New Zealand dataset. As shown in Figure \ref{fig:cv-nz-bar}, the results were highly consistent across folds, with an overall mean SAE of 42.29 km and a median SAE of only 2.95 km. This stability across partitions supports the transfer learning findings, showing that the model’s performance is not dependent on a specific training/test split but generalises reliably to unseen data.

Our results are consistent with recent evaluations of LLM spatial reasoning and spatial cognition. \cite{yang2025evaluating} report that both proprietary and open-source models exhibit limited robustness on spatial tasks that require integrating multiple cues into a globally coherent representation, rather than recalling isolated geographic facts. This aligns with our transfer-learning outcomes: cross-country transfer was minimal without local training data, whereas transfer across independently curated datasets within New Zealand remained comparatively stable, suggesting that the learned mapping is strongly conditioned on regional spatial distributions. Importantly, \cite{yang2025evaluating} further demonstrates that coupling LLMs with deterministic external tools can substantially enhance large-scale spatial cognition, highlighting a promising direction for improving both accuracy and generalisability beyond what text-only inference can achieve.

\begin{table}[]
\tbl{Performance on the New Zealand test set across models fine-tuned with different multi-region training data compositions. Training dataset proportions are relative to the original training data splits (70\%) of each dataset.}{
\begin{tabular}{@{}lllllll@{}}
\toprule
Model      & AUS Training Data             & NZ Training Data              & Mean     & Median   & Within 10km & Within 1km \\ \midrule
GeoLocate  & -                    & -                    & 129.62km & 12.24km  & 10.45\%     & 45.39\%    \\ \midrule
\multirow{6}{*}{Mistral 7B} 
           & 100\% (20,697)        & 0 (0)                & 1314.3km & 1706km   & 2.30\%      & 0.50\%     \\
           & 100\% (20,697)        & 20\% (4,063)          & 57.27km  & 10.39km  & 48.56\%     & 8.10\%     \\
           & 50\% (10,348)         & 50\% (10,158)         & 47.9km   & 5.71km   & 62.83\%     & 15.18\%    \\
           & 0\% (0)              & 100\% (20,318)        & 41.95km  & 3.55km   & 70.43\%     & 25.36\%    \\
           & 20\% (4,139)          & 100\% (20,318)        & \textbf{39km}     & \textbf{3.41km}   & \textbf{70.89\%}     & \textbf{26.07\%}    \\ \bottomrule
\end{tabular}}
\label{tab:transfer-learning-nz-aus}
\end{table}

\begin{table}[b]
\tbl{Analysis of the impact of training dataset size on the accuracy of the fine-tuned LLM using the New Zealand dataset.}
{\begin{tabular}{@{}lllll@{}}
\toprule
Training dataset size & Acc@10km          & Acc@1km         & Median SAE      & Mean SAE         \\ \midrule
0            & 5.23\%           & 0.48\%          & 49.78km         & 1120km          \\
1000                 & 24.42\%          & 2.07\%          & 23.8km             & 70.78km          \\
5000                 & 48.51\%          & 8.16\%          & 10.53km             & 56.4km          \\
10000                 & 60.76\%          & 15.37\%          & 6.15km          & 48.09km          \\
15000                & 67.60\%          & 20.42\%          & 4.31km          & 42.43km \\ \bottomrule
\end{tabular}}
\label{tab:training-size}
\end{table} 

\subsubsection{Summary of Results}

Across all experiments, the fine-tuned Mistral 7B model consistently outperformed baselines when locality descriptions carried richer spatial cues. In smaller/denser (with regard to data) settings, it was strongest (e.g., New Zealand 70.43\% Acc@10 km / 25.36\% Acc@1 km; New York 84.89\% / 66.71\%; Victoria 81.76\% / 22.79\%), and it generalized well to Spanish (Mexico 75.82\% / 51.41\%, median SAE 0.89 km). In larger regions such as the USA and Australia, achieving higher performance required a substantial training dataset for broad geographic coverage. Comparisons with prominent GPT models (GPT-5, GPT-4.1 and GPT-4o) further underscored the value of domain-specific fine-tuning, as general-purpose proprietary models failed to reach similar levels of accuracy.

Content analysis shows what the model actually learned: it integrates multiple toponyms with directional and distance expressions to compose a plausible coordinate, rather than matching a single place name. Longer descriptions, by including more place names and spatial indicators, were associated with lower errors. Moreover, when explicit distances were removed, the fine-tuned LLM’s accuracy declined slightly, suggesting genuine reliance on quantitative cues. Prompt rewordings within the same pattern had only a minor effect, implying that the gains come from learned spatial composition rather than prompt phrasing.

Because performance is closely tied to record density and the geographic extent of the training data, cross-country transfer was limited without local adaptation. Even modest amounts of in-region data yielded measurable improvements, whereas foreign-only training resulted in substantially higher errors. Taken together, these findings show that fine-tuning enables the model to internalize spatial language and compositional cues, achieving strong georeferencing performance across regions and languages when sufficient and representative training data are available.

\subsubsection{Limitations and Future Work}
Although our methodology for georeferencing complex locality descriptions using LLMs yielded favourable results compared to the baselines, we identified several limitations. One key consideration is the need for substantial training datasets. To explore the impact of training dataset size on performance, we fine-tuned the model with varying sizes of the New Zealand dataset. As shown in Table \ref{tab:training-size}, there is a clear trend indicating that the model's performance improves as the training dataset size increases. Notably, these experiments highlight that fine-tuning the LLM with just 5,000 samples is sufficient to surpass the performance of GEOLocate. %This limitation affects the generalizability of our method, given that only a small portion of collection records are currently digitized and georeferenced, and large training datasets with highly reliable ground truth data are scarce.

Another limitation of our LLM-based georeferencing approach is its limited transfer learning capability, since accurate georeferencing often requires local geographic knowledge as discussed in Section~\ref{sec:transfer-learning}. In continuing our research on multi-region fine-tuning, we plan to enhance the model’s ability to access external resources, such as gazetteers and geographic databases, to supplement its georeferencing capabilities. This could be achieved by integrating retrieval-augmented generation (RAG) \citep{guu2020retrieval}, or API-based lookups \citep{schick_toolformer_2023} that allow the model to reference location data dynamically. By incorporating such external knowledge sources, we aim to improve the model’s accuracy in georeferencing data from previously unseen regions and enhance its generalizability across different geographical contexts. 

Additionally, our model currently lacks a measure of the associated uncertainty in the results. This measure is crucial for determining the effectiveness and suitability of the model, an aspect extensively explored in previous studies, such as those by \cite{wieczorek_point-radius_2004, van_erp_georeferencing_2015, guo_georeferencing_2008}. However, several techniques have been developed for estimating the uncertainty of LLM outputs, such as Negative Log-Likelihood (NLL), consistency-based aggregation techniques, or introducing prompting techniques to verbalize confidence scores in LLM outputs \citep{tian2023just, lin2022teaching, lin2023generating}. Furthermore, \cite{catak2024uncertainty} introduce a novel geometric approach that applies convex hull analysis to the spatial embeddings of LLM outputs, quantifying uncertainty based on the dispersion of generated responses. This method leverages the inherent structure of embedding spaces to infer confidence.  The introduction of such spatially-informed Uncertainty Quantification (UQ) techniques for LLMs opens promising directions for our future work. 

\section{Conclusion}
In this paper, we introduced an automatic georeferencing methodology that utilises LLMs. To the best of our knowledge, this study is the first to employ LLMs for georeferencing text data by predicting coordinates that best match the provided locality descriptions. This novel approach utilizes the ability of LLMs to fine-tune for specific domains by leveraging their extensive pre-trained knowledge base and adaptability to new datasets. By integrating domain-specific training data, we enable the model to accurately interpret and align textual descriptions with corresponding geographic coordinates. This fine-tuning process enhances the model's precision in biological collection data georeferencing tasks, ensuring that it can effectively handle the nuances and context-specific terminology unique to different geographic datasets. 

The results show that our approach outperforms commonly used baselines in almost all cases, demonstrating that our methodology is effective in georeferencing complex locality descriptions of biological collection data across various regions and languages compared to existing methods. Due to limited access to the high-performing GPUs, we experimented only with a 7B parameter LLM variant, which still outperformed existing baselines. We believe that it is likely that the results could be further improved with larger LLMs. Utilising models with a larger parameter count could potentially enhance the granularity and accuracy of georeferencing by capturing more intricate patterns and relationships within the data.

Although our model outperformed the baselines, for fully automated georeferencing of biological records, a higher level of accuracy would be desirable when limited training data are available. Additionally, the absence of uncertainty measures for the predicted coordinates is another drawback of our methodology. Nevertheless, this study represents an initial effort to implement a more advanced and comprehensive approach to georeferencing complex locality descriptions in biological collection data using LLMs. Future work will focus on refining the LLM's reasoning processes, with input from georeferencing experts. We also plan to incorporate Retrieval-Augmented Generation (RAG) and integrate external data sources, such as gazetteers and web mapping platforms like OpenStreetMap, to enhance the model’s accuracy and reliability. 

\section*{Disclosure statement}
No potential conflict of interest was reported by the author(s).

\section*{Data and Codes Availability Statement}

The data and codes that support the findings of this study are available at the link \href{https://doi.org/10.6084/m9.figshare.27129759}{https://doi.org/10.6084/m9.figshare.27129759}. 

\section*{Funding}
This work was supported by the Ministry of Business Innovation and Employment Smart Ideas Fund (grant number MAUX2104), New Zealand.

\bibliographystyle{tfv}
\bibliography{biowhere}

\section{Appendices}

\appendix
\section{Sample Responses from LLMs}

\begin{figure}[H]
    \centering
    \fbox{\includegraphics[width=0.70\linewidth]{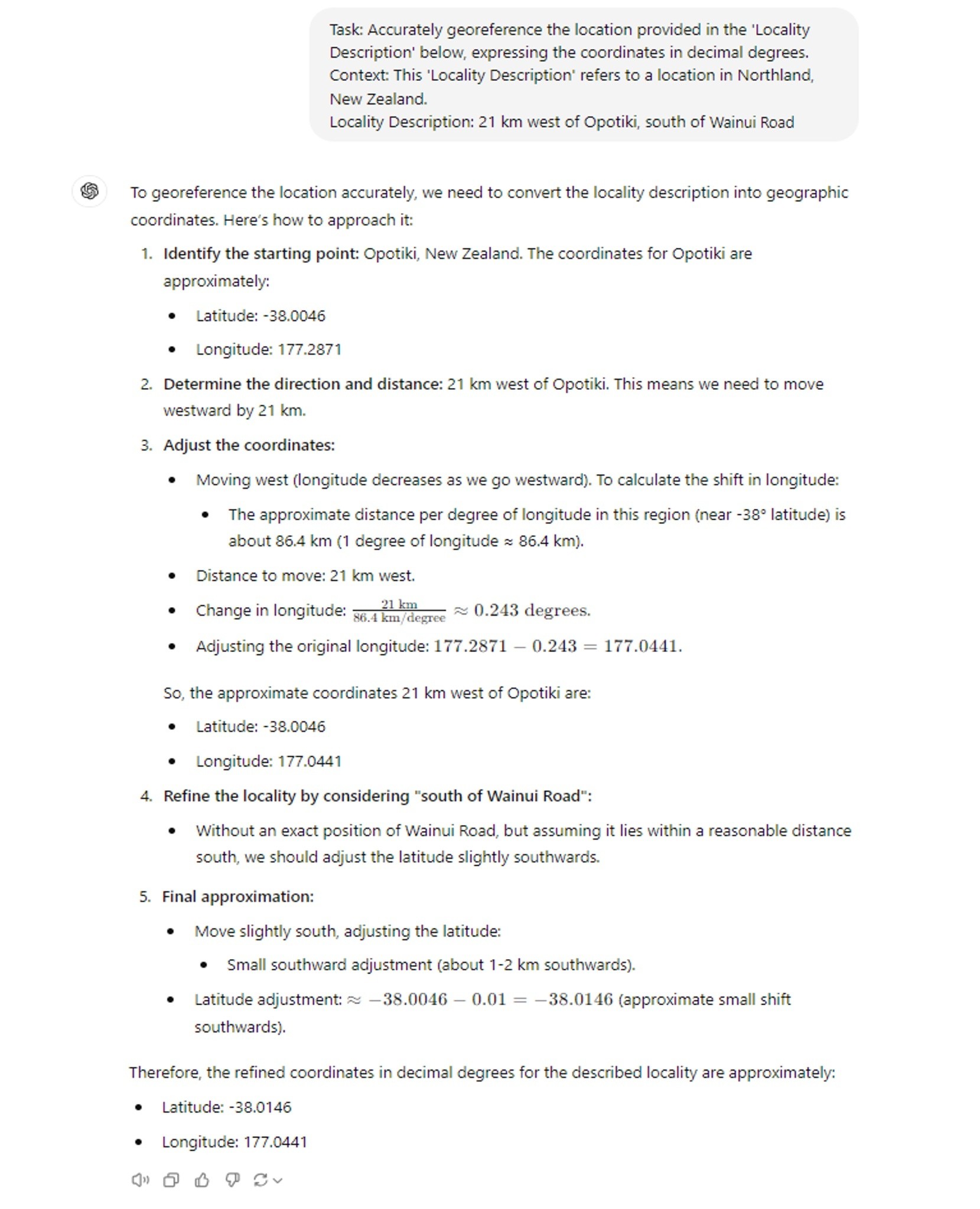}}
    \caption{Response from ChatGPT-4 for a georeferencing problem, generated on June 30, 2024.}
    \label{fig:chatgpt-response}
\end{figure}

The figure \ref{fig:chatgpt-response} shows a response provided by ChatGPT-4 when presented with a georeferencing task. In this example, the context manager pattern was utilized. The response from ChatGPT includes the calculation of offsets from a reference location.

\begin{figure}[H]
    \centering
    \fbox{\includegraphics[width=0.70\linewidth]{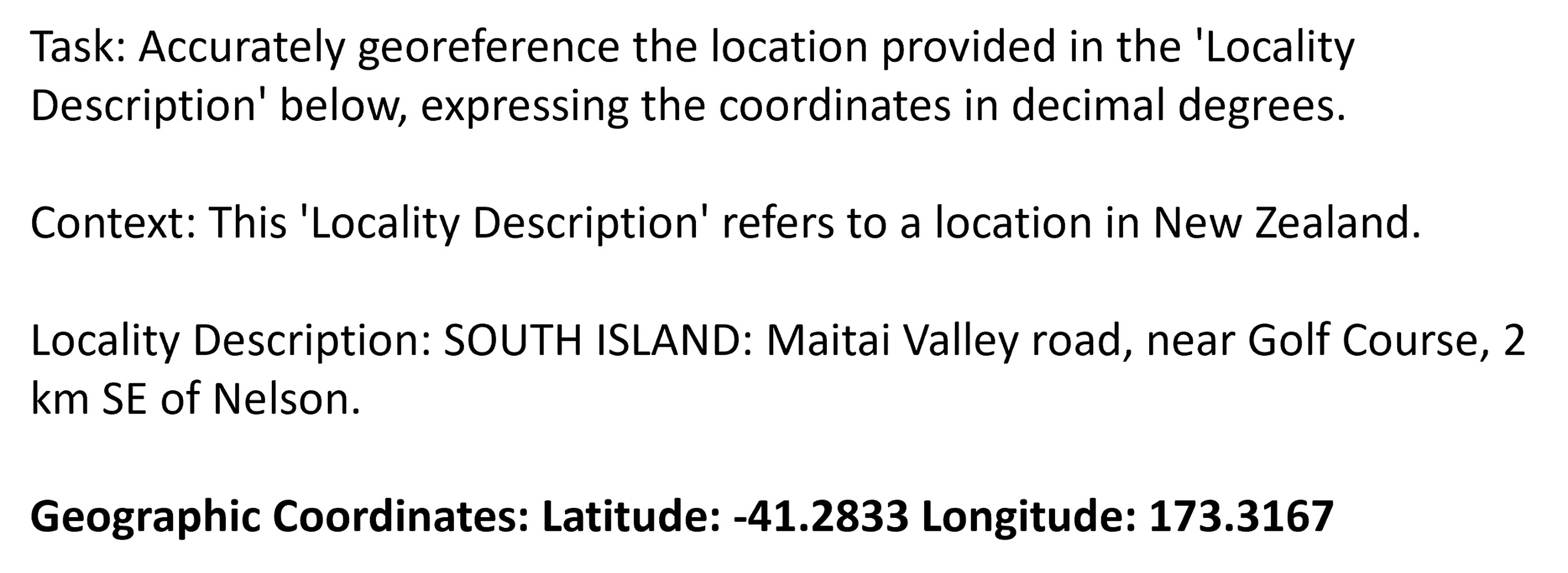}}
    \caption{Response from the fine-tuned mistral model for a georeferencing problem}
    \label{fig:mistral-response}
\end{figure}

Figure \ref{fig:mistral-response} shows how our fine-tuned LLM responded to the task of georeferencing locality descriptions. The response was generated in a manner similar to the formatting of the training data.

\section{Gazetteer-Matching Algorithm}

The following pseudocode summarizes the gazetteer-matching algorithm, which was implemented as one of the baseline methods.

\begin{algorithm}
\caption{Gazetteer-Matching Algorithm}
\begin{algorithmic}[1]
\State \textbf{Input:} localityDescription, state, country
\State \textbf{Output:} latitude, longitude

\State CALL Spacy with localityDescription \textbf{returning} locationEntities
\State possibleLocations $\gets$ empty list

\For{each entity in locationEntities}
    \State CALL GeoNames with entity, state, country \textbf{returning} gazetteerLocations
    \State APPEND gazetteerLocations to possibleLocations
\EndFor

\State CALL DBSCAN with possibleLocations \textbf{returning} clusters
\State clusterWithMostPoints $\gets$ SELECT cluster with the maximum number of points from clusters

\State latitude $\gets$ Average(latitudes of points in clusterWithMostPoints)
\State longitude $\gets$ Average(longitudes of points in clusterWithMostPoints)

\State \Return latitude, longitude
\end{algorithmic}
\end{algorithm}

\end{document}